%% file: 0_main.tex
\pdfoutput=1

\documentclass[11pt]{article}

\usepackage[]{acl}

\usepackage{times}
\usepackage{latexsym}
\usepackage{amsmath}
\usepackage{graphicx}
\usepackage{amsfonts}
\usepackage{booktabs}
\usepackage{multirow}
\usepackage{xurl}
\usepackage{adjustbox}
\usepackage{wrapfig}
\usepackage{tcolorbox}

\usepackage[T1]{fontenc}

\usepackage[utf8]{inputenc}
\usepackage{scalerel,xparse}

\usepackage{microtype}

\title{Aligning Dialogue Agents with Global Feedback \\ via Large Language Model Multimodal Reward Decomposition}

\author{%
    Dong Won Lee$^1$ \quad
    Hae Won Park$^1$ \quad Cynthia Breazeal$^1$ \quad Louis-Philippe Morency$^2$   \\
    MIT$^1$, CMU$^2$\\
    \texttt{dongwonl@mit.edu}
}

\begin{document}
\maketitle
\begin{abstract}

We propose a large language model based reward decomposition framework for aligning dialogue agents using only a single session-level feedback signal. We leverage the reasoning capabilities of a frozen, pretrained large language model (LLM) to infer fine-grained local implicit rewards by decomposing global, session-level feedback. Our first \emph{text-only} variant prompts the LLM to perform reward decomposition using only the dialogue transcript. The second \emph{multimodal} variant incorporates additional behavioral cues, such as pitch, gaze, and facial affect, expressed as natural language descriptions. These inferred turn-level rewards are distilled into a lightweight reward model, which we utilize for RL-based fine-tuning for dialogue generation. We evaluate both text-only and multimodal variants against state-of-the-art reward decomposition methods and demonstrate notable improvements in human evaluations of conversation quality, suggesting that LLMs are strong  reward decomposers that obviate the need for manual reward shaping and granular human feedback.

\end{abstract}

\section{Introduction}
\input{1_intro}

\section{Related Works}

\input{2_related}

\section{Background: Aligning Language Models}
\input{3_problem}

\section{Methods: LLM-GELI}
\label{sec:method}

\input{4_methods}

\input{z97_fig_rewards}
\section{Experiments }

\input{5_experiments}

\section{Results \& Discussion}
\input{6_results}

\section{Conclusion}
\input{8_conclusion}

\section{Limitations}

While LLM-GELI introduces a scalable and interpretable approach to reward decomposition, its effectiveness depends critically on the reasoning capabilities of the underlying large language model. Our method assumes that the pretrained LLM has sufficient social and affective knowledge to infer turn-level credit assignments from global feedback. However, these inferences are ultimately shaped by the LLM’s pretraining data, which may omit domain-specific conversational dynamics or culturally situated behaviors. As such, reward assignments may be inconsistent or misaligned in unfamiliar contexts.

The decomposition procedure is also inherently sensitive to prompt design. Despite using structured prompts grounded in dialogue and behavioral context, small changes in prompt wording or input formatting can lead to significant variation in the resulting reward signals. This introduces instability and reduces reproducibility, particularly when transferring to new domains or LLM backbones. Addressing prompt sensitivity remains an open challenge in prompt-based credit assignment methods.

For the multimodal variant, the quality of decomposition depends on the accuracy of behavioral signal extraction and its subsequent translation into natural language descriptors. While this textual abstraction enables compatibility with language-only LLMs, it may obscure or distort the nuances of the original multimodal data—especially in cases where behavioral cues are ambiguous or conflicting. The resulting reward signals are thus only as reliable as the upstream perception and descriptor generation pipelines.

There are risks that could arise as a result of more social, dialogue agents that can interact with people in a long-term interaction. Conversational agents could be used maliciously for deception, manipulation, and the spread of misinformation. Furthermore, conversational agents which use multimodal data could enhance seriousness of these issues, as models can detect subtle cues such as microexpressions to infer and manipulate the user.

As a potential measure to mitigate such misuse, we plan to release our code and model weights under a license which prevents the use of our assets by any party that support or contribute to false impersonation or hate speech (Do No Harm, Nonviolent Public or Hippocratic License).

\section*{Acknowledgements}
DWL and HWP is supported by the IITP grant funded by the Korean Ministry of Science and ICT (No.2020-0-00842, Development of Cloud Robot Intelligence for Continual Adaptation to User Reactions in Real Service Environments). This research is partially supported by the National Institutes of Health (awards R01MH125740, R01MH132225, and R21MH130767). Any opinions, findings, conclusions, or recommendations expressed in this material are those of the author(s) and do not necessarily reflect the views of the sponsors, and no official endorsement should be inferred. We thank members of the Personal Robots Group at MIT and the Multicomp Lab at CMU for their revisions, feedback and support.


\bibliography{anthology}

\appendix
\label{sec:appendix}
\input{9_appendix}

%


\end{document}

%% file: 1_intro.tex
As dialogue agents become increasingly deployed in open-ended, long-form social interactions, the challenge of aligning model behavior with trajectory-level human preferences becomes more pressing. While large language models (LLMs) have achieved strong linguistic competence, alignment techniques such as reinforcement learning with human feedback (RLHF) \cite{ouyang2022training} remain fundamentally constrained by their reliance on fine-grained, turn-level annotations. In many realistic settings, such as therapy, education, or companionship, annotators provide feedback only at the session level, reflecting their overall impression of the entire interaction. This presents a major obstacle: how can one use sparse, global feedback to improve local behavior?


In this paper, we first introduce \textbf{LLM-GELI}, a  \underline{\textbf{L}}arge-\underline{\textbf{L}}anguage \underline{\textbf{M}}odel based solution to decompose \underline{\textbf{G}}lobal \underline{\textbf{E}}xplicit feedback into \underline{\textbf{L}}ocal \underline{\textbf{I}}mplicit turn-level supervision signals that can be used for RLHF. We leverage a frozen pretrained LLM as a reward decomposition oracle. Given the full dialogue and a scalar global reward, the LLM is prompted to infer how utterances contributed to the final outcome score. The result is a turn-level reward decomposition derived purely from the LLM’s reasoning over the linguistic context. These decomposed signals are then used to train a lightweight, text-only reward function that can guide downstream policy learning.

Furthermore, we introduce \textbf{Multimodal-LLM-GELI}, an extension of our core framework that further incorporates multimodal behavioral signals—such as facial expressions, gaze, and prosody—into the reward decomposition process. Building on recent advances in multimodal prompting, we represent the listener’s nonverbal reactions as natural language descriptors and append them to each utterance in the dialogue history. This textual augmentation enables the language model to interpret affective and social cues in context, allowing it to reason about the listener's internal state and how it evolves throughout the conversation. By grounding decomposition in the listener’s multimodal feedback, Multimodal-LLM-GELI produces turn-level reward estimates that more accurately reflect the listener's extralinguistic reactions. 

We evaluate our approach on the CANDOR dataset \cite{reece2023candor}, which features long-form video conversations and post-session survey ratings, and an out-of-distribution dataset, SODA \cite{kim-etal-2023-soda}. Our results show that LLM-GELI and Multimodal-LLM-GELI outperform prior baselines across a range of human and automatic metrics. Notably, Multimodal-LLM-GELI achieves the strongest gains on affective and socially grounded human evaluation measures \cite{lee2022evaluating}, which highlights the value of LLMs not only as generators, but as competent interpreters of both global human preferences and rich, real-world behavioral signals to serve as reward decomposition oracles.

%% file: 2_related.tex
\input{z96_fig_overview}

\paragraph{Multimodal Prompting} Recent advances in multimodal learning highlight the growing potential of large language models (LLMs) as universal interfaces for integrating and reasoning across vision, language, audio, and sensor modalities. Socratic Models \cite{zeng2022socratic} introduced a modular framework for zero-shot multimodal reasoning by composing frozen unimodal models—such as CLIP \cite{radford2021learning} and GPT-3—through language-based hypothesis exchange, eliminating the need for joint training. Building on this paradigm, MM-ReAct \cite{yang2023mm} leverages ChatGPT to orchestrate visual experts (e.g., OCR, object detectors) via prompt-based planning. Furthermoe, recent efforts have explored using LLMs for time-series and sensor data through language interfaces. Health-LLM \cite{kim2023aligning} and PHIA \cite{merrill2024transforming} convert wearable sensor data into enriched natural language prompts, enabling personalized health prediction and recommendation in a zero-shot setting. Together, these works illustrate a growing shift toward using language as a medium for zero-shot reasoning with multimodal behavioral data.

\vspace{-0.08in}
\paragraph{Reward Decomposition in Reinforcement Learning} Reward Decomposition, also known as Temporal Credit Assignment (TCA), is a fundamental concept in reinforcement learning that addresses the challenge of determining how past actions contribute to future outcomes. Effective TCA enables an agent to properly attribute delayed rewards to the decisions that caused them, thus improving learning efficiency and policy quality. One classical approach to TCA involves manipulating the $\lambda$-return in temporal-difference learning, which adjusts the balance between short-term and long-term credit assignment \citep{petrik2008biasing, jiang2015dependence}. More recent advances have reformulated TCA as a return decomposition problem. For instance, RUDDER \citep{arjona2019rudder} redistributes rewards by measuring the predictive information gained between consecutive states, allowing for more precise step-wise credit assignment. IRCR \citep{gangwani2020learning} implements a uniform redistribution scheme, while Randomized Return Decomposition (RRD) \citep{ren2021learning} frames the problem as a Monte Carlo regression task, using least-squares estimation to infer per-step rewards from trajectory-level returns.

\paragraph{LLMs for Reward Decomposition} Recent work has explored using large language models (LLMs) to support reward decomposition in reinforcement learning (RL). For instance, LaRe \cite{qu2025latent} prompts an LLM to generate a latent code that is then decoded into scalar rewards which sum to the total episodic return. CALM \cite{pignatelli2024assessing} employs an LLM to identify when subgoals are achieved, assigning auxiliary rewards in a zero-shot manner. In contrast, recent alignment methods like SPIN \cite{chen2024self} and \cite{go2023compositional} treat LLMs as reward functions, using them to rank or evaluate outputs via self-play or preference modeling—requiring dense or per-turn supervision. On the other-hand, GELI \cite{lee2024global} is closely related to reward decomposition for dialogue modelling, however, it does not utilize an LLM, instead it utilizes a manually designed reward shaping function. Our method, LLM-GELI, is the first to leverage a pretrained LLM that decomposes a global end-of-dialogue reward into fine-grained, turn-level signals. Furthermore, by prompting an LLM with multimodal behavioral feedback (e.g., facial affect, prosody), we infer fine-grained turn-level rewards from sparse, trajectory-level supervision.




%% file: z96_fig_overview.tex
\begin{figure}[t]
    \centering  
    \includegraphics[width=1.05\columnwidth]{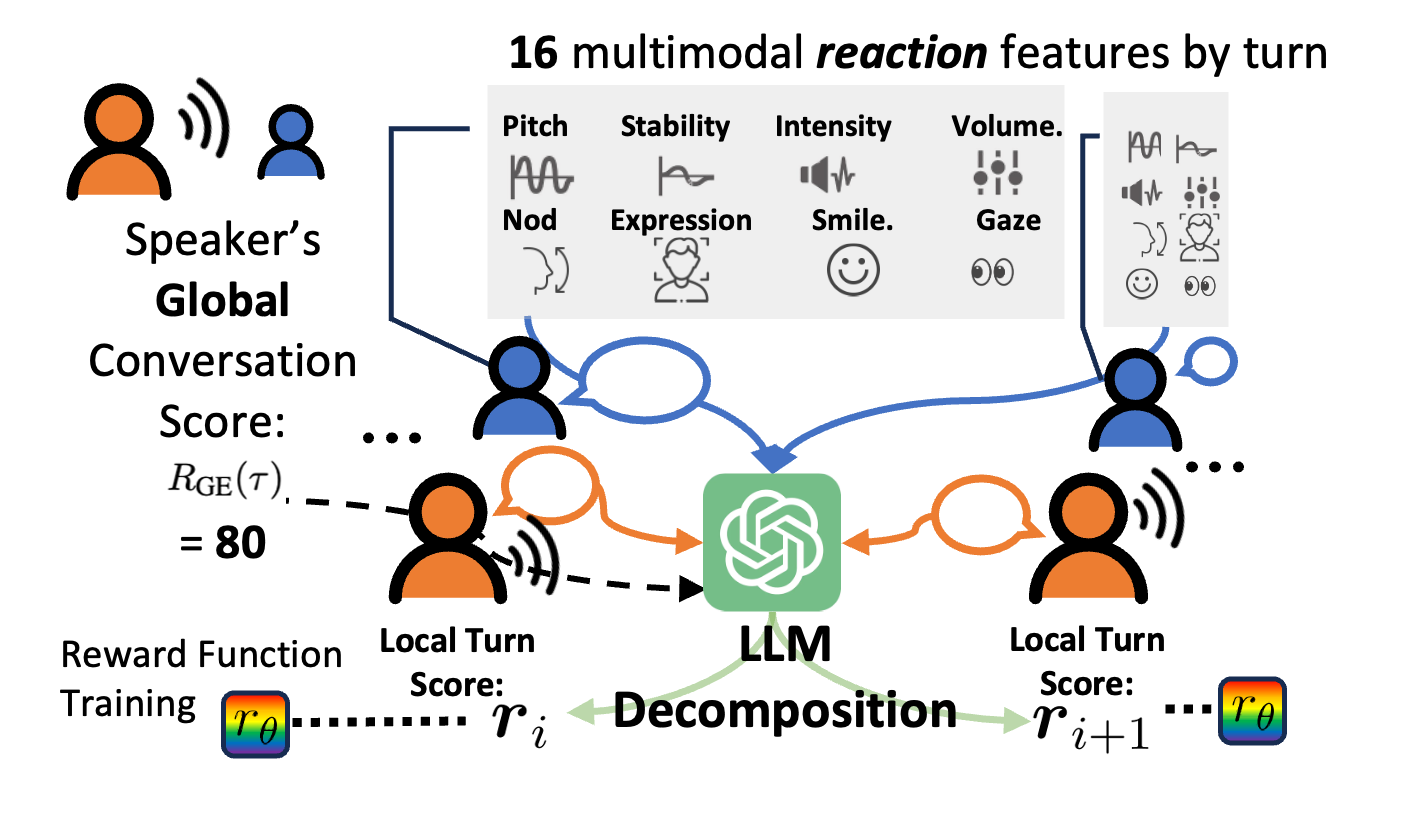}
    \caption{Multimodal LLM-GELI framework. A global reward $R_{\text{GE}}(\tau)$ is decomposed into turn-level pseudo-rewards using an LLM prompted with dialogue and multimodal listener cues. These local rewards supervise a text-only reward model $r_\theta$, distilling crossmodal reasoning into a learnable function for alignment (Sec. \ref{sec:method}).}
    \label{fig:overview}
    \vspace{-5mm}
\end{figure}


%% file: 3_problem.tex
We consider the problem of generating conversational responses using an autoregressive language model in a multi-sensory setting. We model the language generator as a dialogue agent with a policy $\pi_\phi$, parameterized by $\phi$, following prior work \cite{liu2018dialogue, liang2020moss, wen2016network, thoppilan2022lamda}. At each turn $t$, the agent observes the dialogue history $s_t$—which includes all preceding utterances $s_1, a_1, \dots, s_{t-1}, a_{t-1}$—and produces a response $a_t$ as its action. For notational simplicity, we denote the full dialogue context up to turn $t$ as $s_t = s_{\left[:t-1\right]}$. The policy $\pi_{\phi}(s_t)$ thus defines a distribution over possible responses $a_t$ conditioned on $s_t$.

To align the agent’s behavior with human preferences, reinforcement learning from human feedback (RLHF) is commonly employed \cite{ouyang2022training, jaques2020human, stiennon2020learning}. Given a reward function $r_\theta(s_t, a_t)$, parameterized by $\theta$, that scores the quality of a generated response in context, we optimize the policy via reinforcement learning. A regularization term based on the Kullback–Leibler (KL) divergence from the original pretrained model is typically included to prevent the policy from straying too far from its initialization. The resulting objective takes the form:

{\small
\begin{equation}
\max_{\phi} \,\, \mathbb{E}[r_\theta\left(s_t, a_t\right)] - \gamma D_{KL}(\pi_{\phi}\left(\cdot |s_t\right)||\pi_{\eta}\left(\cdot |s_t\right)),
\end{equation}
}

where $\pi_\eta$ is a reference model.




%% file: 4_methods.tex
In many long-form dialogue settings, we often lack fine-grained annotations for individual utterances. Instead, we may only have access to a single \textit{global explicit} (GE) reward, a scalar score assigned at the end of an interaction that reflects the overall impression or outcome, such as how positively the user felt about the conversation. A key challenge is how to decompose this global reward into per-turn feedback that can train a reward model capable of assigning appropriate values to individual utterances. We address this challenge using a \textit{Large Language Model GELI (LLM-GELI)} framework, which proceeds in two stages (1) Global Explicit (GE) Reward Decomposition using the zero-shot prompting  capabilities of a pretrained LLM and (2) Local Implicit (LI) Reward Modeling.

\subsection{Global Explicit Reward Decomposition via Zero-shot Prompting of LLM
} \label{sec:ge}
We consider a long-form dialogue trajectory 
\[
\tau = \{(s_0, a_0), (s_1, a_1), \dots, (s_T, a_T)\},
\]
where \(s_t\) is the dialogue state (e.g., conversational history) and \(a_t\) is the agent's utterance at turn \(t\). At the end of the episode, we observe a scalar \textit{global explicit reward} \(R_{\text{GE}}(\tau) \in \mathbb{R}\), which reflects an overall human judgment of the interaction (e.g., how positively the user felt about the conversation).

Instead of assuming access to fine-grained reward supervision at each turn, we utilize a \textit{LLM} as a reward decomposition oracle. The LLM \(\mathcal{M}\) is prompted with the full dialogue \(\tau\), the global reward \(R_{\text{GE}}(\tau)\), and a set of the listener's aligned multimodal feedback features \(\{x_t\}_{t=0}^{T}\), such as facial expressions or prosody. 

The Multimodal-LLM-GELI outputs a turn-level reward decomposition that utilizes multimodal features and the dialogue transcript:

{\small
\begin{equation}
\mathcal{M}_{\text{MM-LLM-GELI}}\left(\tau, \{x_t\}_{t=0}^{T}, R_{\text{GE}}(\tau)\right) = \left\{ r_{\text{LI}}(s_t, a_t) \right\}_{t=0}^{T},
\end{equation}
}

The LLM-GELI outputs a turn-level reward decomposition based only on the dialogue transcript:
\begin{equation}
\mathcal{M}_{\text{LLM-GELI}}\left(\tau, R_{\text{GE}}(\tau)\right) = \left\{ r_{\text{LI}}(s_t, a_t) \right\}_{t=0}^{T},
\end{equation}

\vspace{-2mm}
subject to the (\emph{soft}) constraint in the prompt:
\begin{equation}
\sum_{t=0}^{T} r_{\text{LI}}(s_t, a_t) = R_{\text{GE}}(\tau).
\end{equation}
\vspace{-4mm}

In practice, this corresponds to the following prompt given to the LLM:

\begin{tcolorbox}
\textbf{Prompt:}
You are given a final scalar score for Speaker A, indicating how much Speaker B experienced positive feelings (e.g., good, pleasant, happy) or negative feelings (e.g., bad, unpleasant, unhappy) during the conversation. 

\vspace{1mm}
Final score: \{\emph{Session Score} ($R_{\text{GE}}(\tau)$)\}
Below, you are given the full conversation with aligned multimodal conversational features. Consider the utterances and the features, focusing on the utterances of Speaker A and the reactions of Speaker B. Redistribute the score across each of Speaker A’s utterances.

 Whenever you identify a salient, important utterance that contributed positively or negatively to the final reward score, assign a numerical value (which may be positive, negative, or zero) representing its contribution. 
\vspace{1mm}
All of the assigned scores must sum up to:\{\emph{Session Score} ($R_{\text{GE}}(\tau)$)\}.

\vspace{1mm}
\{\emph{Transcript / Multimodal Features}\}
\end{tcolorbox}

This decomposition provides a \emph{local implicit} reward signals \(r_{\text{LI}}\) that reflects the LLM's judgment of each action's contribution to the final reward. Importantly, these targets encode semantic, behavioral, and extralinguistic information (in the multimodal variant), as the model reasons over responses and reactions (e.g., listener facial affect). In our experiments, we use \texttt{o3-mini} \cite{openai2024o3mini} as the backbone reward decomposition LLM, \(\mathcal{M}\). 

\subsection{Reward Modeling with LLM-Derived Local Implicit Rewards}

Once turn-level rewards are obtained from the LLM, they are treated as \textit{local implicit} (LI) supervision signals, $r_{\text{LI}}$, and used to train a lightweight reward function \(r_{\theta}\), which conditions only on the textual dialogue context, i.e., state-action pairs \((s_t, a_t)\). Although the Multimodal-LLM-GELI had access to aligned multimodal cues during decomposition, the reward model $r_\theta$ does not. Instead, it learns from the LLM derived pseudo-labels, effectively distilling high-level, crossmodal reasoning into a purely text-conditioned reward function. This reward function is optimized via a standard mean squared error (MSE) loss:

\begin{equation}
L_{\text{LI}}(\theta) = \mathbb{E}_{(s_t, a_t) \sim \mathcal{D}} \left[ \left( r_{\text{LI}}(s_t, a_t) - r_{\theta}(s_t, a_t) \right)^2 \right],
\end{equation}

where \(r_{\text{LI}}(s_t, a_t)\) denotes the pseudo-ground-truth reward for turn \(t\) obtained from the LLM-GELI method, and \(r_{\theta}(s_t, a_t)\) is the learnable text-only reward function parameterized by \(\theta\).

%% file: z97_fig_rewards.tex
\begin{figure*}[t]
    \begin{center}
    \includegraphics[width=1\textwidth]{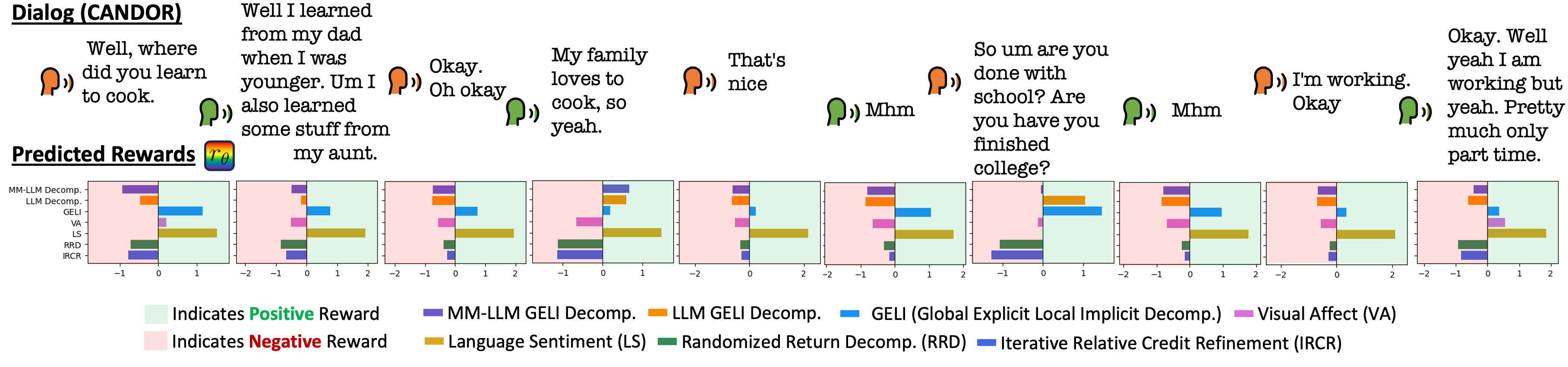}
    \caption{Comparison of reward decomposition across models on a CANDOR dialogue segment. MM-LLM GELI (purple) assigns contextually appropriate, affect-sensitive local rewards, aligning closely with social-emotional cues, better grounding in multimodal feedback, and stronger alignment with human intuition. See Sec. \ref{sec:decomposed}.}
    
    \label{fig:reward}
    \end{center}
    \vspace{-7mm}
\end{figure*}

%% file: 5_experiments.tex
In this section, we evaluate the effectiveness of the proposed \textsc{LLM-GELI} framework for preference-based learning, namely reward modeling and language model adaptation in long-form dialogue. Therefore, our evaluation consists of two stages. First, we train reward functions using the various methods of global explicit reward decomposition, which we denote as ``\textbf{GE}'', and/or local implicit reward supervision, which we denote as ``\textbf{LI}''. Second, we fine-tune a dialogue language model using Proximal Policy Optimization (PPO) \cite{schulman2017proximal}, guided by the learned reward functions to improve response generation. Due to computational restraints, all reward function training and RL fine-tuning procedures are executed with a single run. We assess each method’s impact using both automatic and human-centric metrics.

\subsection{Dataset}
We conduct our training and evaluation experiments using the \emph{CANDOR} dataset \cite{reece2023candor}, chosen for its long-form, naturalistic dialogue structure—featuring an average of 159.4 turns, 31.3 minutes per conversation, and 17.81 words per turn—along with its scale (1,656 conversations, over 7 million words, and 850 hours of data). CANDOR also includes aligned video recordings, enabling multimodal analysis common in face-to-face conversation corpora. We use this dataset both to train reward functions and to sample dialogue contexts for response generation. Specifically, we construct disjoint splits: a held-out set of approximately 30,000 dialogue history–utterance pairs for reward model training, and a larger set of roughly 100,000 pairs for reinforcement learning. The global explicit (GE) signal is derived from the post-interaction survey item: “Overall during your conversation, to what extent did you feel positive feelings (e.g., good, pleasant, happy) or negative feelings (e.g., bad, unpleasant, unhappy)?”. 
To assess generalization, we also evaluate on the \emph{SODA} dataset \cite{kim-etal-2023-soda}, a large-scale synthetic social dialogue corpus generated via GPT-3.5 and grounded in a structured social commonsense knowledge graph. Human evaluations indicate that SODA conversations are more consistent, natural, and specific than other human-authored datasets. It contains 1.5 million dialogues, with an average of 7.6 turns and 16.1 words per turn.

\paragraph{Multimodal Features} We utilize a set of 17 multimodal conversational features capturing prosodic, visual-social, and linguistic cues to model the affective dynamics of interactions. Audio-based prosodic features include (1) \texttt{f0} (fundamental frequency, representing pitch) calculated using the parselmouth library \cite{jadoul2018introducing},  (2) \texttt{intensity} (voice amplitude) from a predictor trained on the RAVDESS dataset \cite{livingstone2018ryerson}, (3)\texttt{jitter} (indicative of pitch instability from disvoice library \cite{dehak2007modeling}, and (4) \texttt{log energy} (signal power, capturing loudness) from librsoa library \cite{mcfee2020librosa}. Visual-social features extracted from the video modality include detectors such as (5) \texttt{gaze} (whether the speaker is looking on-screen) calculated based on white pixel value counts, (6) \texttt{nod\_yes} , (7) \texttt{nod\_no}(indicating head nodding behaviors) where a nod is defined as having a nod zero crossing threshold >= 2 and a nod scaled face range threshold > 0.1 , and (8) \texttt{smile} (presence of a smile) from a smile detector (cv2.CascadeClassifier) \cite{opencv_library} . Additionally, we incorporate emotion recognition scores derived from facial expression classifiers trained on AffectNet \cite{mollahosseini2017affectnet}, including probabilistic scores for (9 - 16) \texttt{anger, contempt, disgust, fear, happiness, neutral, sadness, and surprise}. Furthermore, we also include the linguistic content, the (17) \texttt{utterance} of the speakers.

\paragraph{Language Descriptors} To enable effective integration with language models, we convert continuous multimodal behavioral features, which vary over time, into discrete natural language descriptors aligned with each utterance. Since these features are sampled at a different temporal resolutions than utterances (e.g., per frame or per second), we aggregate the feature values over the utterance duration by computing their mean and appending them to the corresponding utterances. This formulation allows us to leverage the zero-shot multimodal reasoning capabilities of large language models, as outlined in prior work on language-as-interface paradigms. An example of such processed sample is included in the Supplementary. 

\paragraph{Focusing on Listener Reactions} Since our goal is to decompose a global reward based on the listener’s affective response, we condition the model exclusively on the multimodal features of the listener (i.e., the participant who completed the post-interaction survey and provided the overall affective rating). This choice centers the reward attribution on the reactions and experiences of the evaluator rather than the speaker, naturally aligning with the perspective-taking required for accurate reward decomposition.

\subsection{Baseline Models}
\input{z82_compare}
We compare LLM-GELI against several state-of-the-art Glbal Explicit reward decomposition methods designed to learn fine-grained a reward function from a single global explicit (GE) annotation. To ensure fair comparison, we also evaluate models trained using only local implicit (LI) multimodal signals. All methods build on a small BART model~\cite{lewis2019bart} fine-tuned for conversational summarization,\footnote{\url{https://huggingface.co/kabita-choudhary/finetuned-bart-for-conversation-summary}} with additional linear layers trained for reward prediction. Below, we describe the methods used for comparison. In Figure \ref{fig:compare}, we showcase the differences of various decomposition methods.

\textbf{GE: RRD (Randomized Return Decomposition)} \cite{ren2021learning} reformulates return decomposition using Monte Carlo sampling to scale least-squares reward redistribution to long-horizon tasks. \textbf{GE: IRCR (Iterative Relative Credit Refinement)} \cite{gangwani2020learning} implements uniform reward redistribution by assigning normalized trajectory scores as transition-level proxy rewards.\textbf{GE: RUDDER (Return Decomposition for Delayed Rewards)} \cite{arjona2019rudder} uses an LSTM-based return predictor to assign step-level credit via differences in consecutive return estimates. \textbf{LI: Visual Affect (VA)} leverages facial expressions, via a CNN-based affect classifier trained on AffectNet \cite{mollahosseini2017affectnet}, as local rewards. \textbf{LI: Language Sentiment (LS)} models implicit reward from utterance sentiment using a pretrained multilingual DeBERTa classifier \cite{he2020deberta, pang2023leveraging}. \textbf{GE \& LI: Global Explicit Local Implicit Decomposition (GELI)} \cite{lee2024global} jointly trains a reward function to redistribute a global explicit reward using RRD \cite{ren2021learning}, shaped by local implicit cues, such as facial affect or gaze. However, their method required carefully hand-engineered functions to interpret these nonverbal signals into local implicit rewards.

\paragraph{Evaluation of Reward Functions.}
To assess the quality of learned reward functions, consistent with prior work \cite{lee2024global}, we report two key metrics: the Global Loss and the Local Difference. The Global Loss, $\mathcal{L}_\text{GE}(\theta)$, quantifies how accurately the predicted fine-grained rewards sum to the global explicit reward $R_\text{GE}(\tau)$ across a dialogue trajectory $\tau$. It is computed as the mean squared error (MSE):

{\small
\begin{align} \label{eq:return_decomposition_loss}
	\mathcal{L}_{\text{GE}}(\theta) = \mathop{\mathbb{E}}_{\tau\sim\mathcal{D}} \left[\biggl(R_{\text{GE}}(\tau)-\sum_{t=0}^{T-1} r_{\theta}(s_t, a_t)\biggr)^2\right]
\end{align}
}

To evaluate alignment with local implicit signals, we introduce the Local Difference, defined as the difference in expected predicted rewards conditioned on positive versus non-positive facial affect (as detected from visual feedback). Formally:

{\small
\begin{equation}
\begin{split}
  \Delta \hat{r}_{LI} & = \mathbb{E}\left[r_\theta(s_t,a_t)|f(s^{mm}_{a_t}) = \textit{positive affect} \right] \\ 
  & -  \mathbb{E}\left[r_\theta(s_t,a_t)|f(s^{mm}_{a_t}) \neq \textit{positive affect} \right]
\end{split}
\end{equation}
}

This metric captures how the predicted reward responds to visual cues from the listener. Under the assumption that positive visual affect implies that the user has received the utterance positively, we expect $\Delta \hat{r}_{\text{LI}} > 0$, indicating alignment between the reward signal and implicit user visual feedback from facial expression.

\subsection{Fine-Tuning Language Models via Reinforcement Learning}

We adapt the LLaMA-2 model~\cite{touvron2023llama} \footnote{We selected this model to mitigate potential confounds arising from data leakage, as the training data of more recent LLaMA variants may include  the CANDOR dataset (public release date: March 2023).} using reinforcement learning with feedback from the trained reward functions. Specifically, we employ the PPO-based RLHF framework from the TRL library~\cite{vonwerra2022trl}, with LoRA~\cite{hu2021lora} applied to reduce memory and compute overhead. The complete hyperparameter settings are provided in Appendix \ref{app:hyper}.

\paragraph{Evaluation of Generated Responses.}
We conduct a human evaluation following prior protocols~\cite{lee2022evaluating,lee2024global}, using 9  criteria commonly used to assess dialogue quality. A total of 300 crowd workers were recruited via Amazon Mechanical Turk. Annotators were presented with dialogue histories and model-generated responses, and asked to identify which model best satisfied each criterion. At the end of the task, participants were asked which model they would prefer to converse with again. This is described in Section \ref{human_eval}.

%% file: z82_compare.tex
\begin{figure*}[!htb]
    \begin{center}
    \includegraphics[width=1\textwidth]{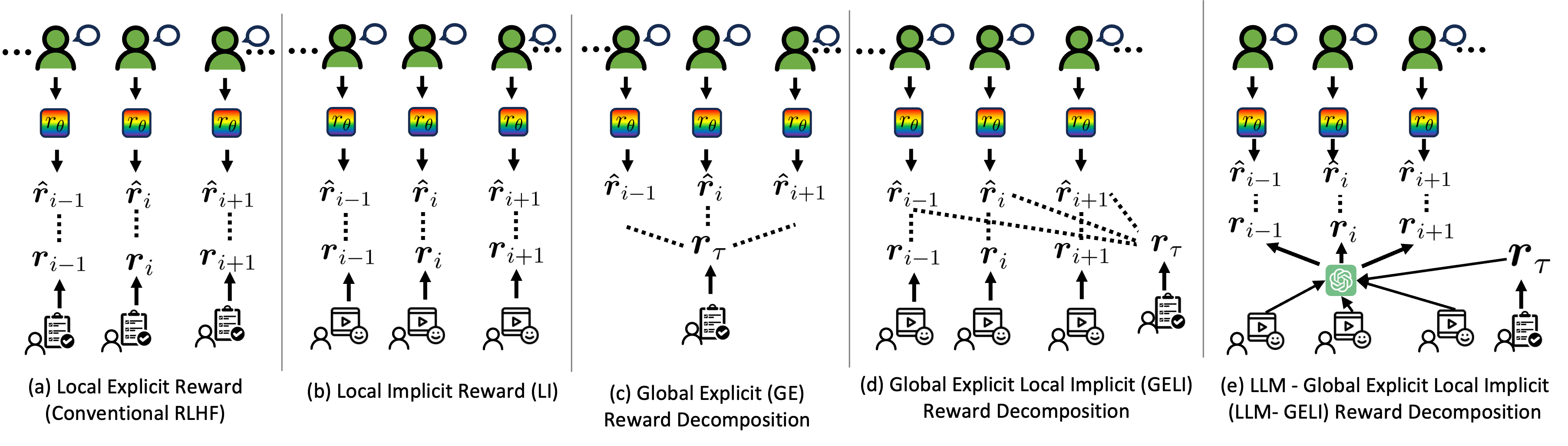}
    \caption{High level schematic of different Reward Modeling and Reward Decomposition Methods. Arrows (→) represent forward passes or assignments (- - -) indicate supervised training targets used to train the reward model ${r_\theta}(\cdot) = \hat{r} \approx r$.
    (a) Local Explicit Reward Modelling, used in conventional RLHF \cite{ouyang2022training}, (b) Local Implicit Rewards where signals are derived from proxy rewards in the form of outputs from a trained task-relevant function which takes in multimodal or task-related signals and outputs a score, \cite{pang2023leveraging} (c) Global Explicit Reward Decomposition, where there is a single trajectory level reward and is decomposed to allocated scores for each step. \cite{arjona2019rudder, ren2021learning, gangwani2020learning} (d) GELI \cite{lee2024global} approach where the global explicit rewards and local implicit rewards are used in conjunction. (e) Our proposed LLM-GELI approach where the reward decomposition is offloaded to a pretrained LLM, which interprets the local listener reaction cues.}
    \label{fig:compare}
    \end{center}
    \vspace{-5mm}

\end{figure*}

%% file: 6_results.tex
\input{z83_tab_candor}

In this section, we discuss the quantitative results and human evaluation of our experiments. 


\vspace{-2mm}
\subsection{Human Evaluation}
\label{human_eval}
Table \ref{tab:human_eval} presents human evaluation results on an unseen splits of the CANDOR dataset. Our approach is the best-performing method in 6 out of 9 metrics. The Multimodal-LLM-GELI model—by augmenting each utterance with multimodal descriptors (e.g., facial affect, gaze)—further improves performance, achieving the best scores in \textbf{Social} (26.33 ± 1.25, $p < 0.01$) \textbf{Interestingness} (21.67 ± 1.89, $p < 0.01$), \textbf{Specificity} (89.67 ± 1.25, $p < 0.05$), and \textbf{Sensibility} (94.67 ± 2.49), \textbf{Fluency} (93.33 ± 0.94, $p < 0.05$). The LLM-GELI model achieves  the best scores in \textbf{Inclination} (25.33 ± 3.77, $p < 0.01$). Furhtermore, our proposed approach results in staticially significant improvements over Base-Llama in 7 out of 9 metrics. For the SODA \cite{kim-etal-2023-soda} unseen out-of distribution dataset, as shown in Table \ref{tab:soda}, Multimodal-LLM-GELI adaptation in staticially significant improvements over Base-Llama in 6 out of 9 metrics, and performing the best for 2 out of 9 metrics \textbf{Connection} (7.67 ± 1.25, $p < 0.01$) and \textbf{Positivity} (5.67 ± 0.9, $p < 0.01$). We further evaluate on ESConv \cite{liu2021towards}, a real-world human-human emotional support corpus, where we find that LLM-GELI and Multimodal-LLM-GELI outperform baselines across most (5 out of 9) human-centric evaluation metrics, including \textbf{Positivity} (36.03 ± 2.88, $p < 0.01$), \textbf{Social} (34.94 ± 4.58, $p < 0.01$), \textbf{Inclination} (38.95 ± 4.42*, $p < 0.01$), \textbf{Interestingness} (1.04 ± 4.07*, $p < 0.01$) and \textbf{Reuse} (9.93 ± 5.04, $p < 0.01$), which are key dimensions for emotionally supportive dialogue. This further confirms the external validity and transferability of our method to real-world social settings like therapy or emotional support conversations.

These strong results of LLM-GELI and Multimodal-LLM-GELI, stands in stark contrast to GELI, which requires a carefully designed manual reward shaping function. This indicates that LLMs can serve as effective decomposers of sparse global feedback into reward signals, bypassing the need for task-specific engineering and domain heuristics. This dramatically reduces the overhead in reward shaping function design, while enabling scalable alignment of agents with nuanced human preferences. We refer readers to Appendix Sec. \ref{app: generations} for real utterance generations of Multimodal-LLM-GELI and LLM-GELI compared to previous baselines.

\subsection{Reward Function Evaluation}




\input{z88_small_tab_rewards}
\paragraph{Global Loss ($L_{GE}$):}
Table \ref{tab:small_reward} (left) reports the Global Loss $L_{GE}$, which quantifies how well the predicted local rewards sum to the ground-truth global feedback, as defined in Eq.~\ref{eq:return_decomposition_loss}. While traditional return decomposition methods such as RUDDER and IRCR achieve moderate MSE scores, our proposed \textbf{LLM GELI} methods—particularly the \emph{Multimodal (MM-LLM)} variant—achieve the \emph{lowest} global loss by a large margin. Notably, this is achieved without any hand-crafted return redistribution, in contrast to methods like GELI, and methods are explicitly optimized to minimize $L_{GE}$ like RUDDER and RRD. This result highlights that leveraging local implicit feedback from pretrained LLMs (and multimodal inputs) can lead to more accurate reward decomposition. The low $L_{GE}$ of both the Language-Only and Multimodal LLM GELI models suggests that LLM-based redistribution offers a generalizable and domain-agnostic solution.

\paragraph{Local Difference ($\Delta\hat{r}_{LI}$):}
Table~\ref{tab:small_reward} (right) shows the Local Difference metric $\Delta\hat{r}_{LI}$, which measures the alignment of predicted reward differences with local implicit feedback—in our case, the difference in predicted reward for turns with positive versus negative visual affect. Models trained solely with global feedback (GE-only methods) yield near-zero $\Delta\hat{r}_{LI}$, indicating their inability to distinguish between positive and negative affective states. In contrast, LI-only baselines that directly optimize on multimodal inputs such as visual affect (VA) yield much higher $\Delta\hat{r}_{LI}$, validating the utility of these local signals.

We observe that the LLM-GELI-(Multimodal) model achieves a non-trivial $\Delta\hat{r}_{LI}$ (0.012), outperforming all GE baselines and approaching the behavior of LI-tuned models. This suggests that the implicit structure captured by the LLM reward decomposition can encode meaningful multimodal cues, even without explicitly training on them. In contrast, the Language-Only LLM GELI variant does not show this sensitivity, reinforcing the importance of multimodal inputs for capturing socially grounded implicit signals. Together, these findings demonstrate that LLM GELI not only eliminates the need for hand-designed reward functions but also captures affectively salient local distinctions when multimodal inputs are present.

\subsection{Comparison of Decomposed Rewards} 
\label{sec:decomposed}

Figure~\ref{fig:reward} presents a turn-by-turn comparison of reward decompositions from different models on a CANDOR dialogue segment. Across the conversation, we observe that \textbf{MM-LLM GELI} (purple) consistently aligns rewards with socially meaningful content (e.g., self-disclosures, affirmations), while discounting low-affect or filler utterances like “Mhm” or “Okay.” In contrast, baseline methods like GELI (blue), VA (pink), and LS (orange) often misattribute reward to neutral or socially uninformative turns, or fail to identify affective shifts. The RRD and IRCR models, optimized for global loss, show erratic or flat credit assignment.

The key difference between \textbf{MM-LLM Decomp.} and \textbf{LLM Decomp.} lies in multimodal grounding: MM-LLM uses local visual affect (e.g., facial expressions, gaze) to inform reward assignments, allowing it to distinguish between utterances with similar surface forms but divergent social impacts—something the language-only variant cannot reliably capture. In the segment where Speaker A asks, “So um are you done with school? have you finished college?”, we observe a key divergence in how models assigned rewards. Both GELI and LLM Decomp. attribute either neutral or slightly positive reward, likely based solely on the linguistic surface (a question showing interest). However, MM-LLM Decomp. assigns a clearly negative reward. This contrast arises because MM-LLM incorporates multimodal cues—such as facial expression, vocal hesitancy, or disengaged affect from the listener—which may signal discomfort or a negative emotional reaction to the question.

%% file: z83_tab_candor.tex
\begin{table*}[t]
\centering
\resizebox{\textwidth}{!}{
\begin{tabular}{@{}lccccccccc@{}}
\toprule
CANDOR \cite{reece2023candor} & Connection & Positivity & Social & Inclination & Interestingness & Reuse & Specific & Sensible & Fluency \\
 & \multicolumn{9}{c}{(/100\%) ↑} \\ \midrule
Human & 6.33 ± 2.05 & 7.67 ± 1.25 & 7.33 ± 2.49 & 7.0 ± 2.16 & 8.67 ± 0.94 & 8.0 ± 1.63 & 76.0 ± 0.82 & 78.67 ± 1.7 & 79.0 ± 2.94 \\
Llama & 20.67 ± 3.77 & 22.0 ± 3.27 & 23.67 ± 3.09 & 19.0 ± 2.83 & 8.67 ± 2.49 & 16.33 ± 4.71 & 86.33 ± 2.36 & 94.33 ± 2.49 & 92.33 ± 3.3 \\
Llama + GE: RRD & 2.67 ± 1.25 & 2.0 ± 0.82 & 2.67 ± 0.47 & 2.67 ± 0.94 & 2.67 ± 0.94 & 2.33 ± 0.94 & 71.67 ± 1.25 & 69.0 ± 0.82 & 69.33 ± 3.77 \\
Llama + LI: LS & 16.67 ± 1.25 & 15.33 ± 1.25 & 18.0 ± 1.41 & 13.0 ± 2.16 & 6.33 ± 0.47 & 11.33 ± 1.7 & 86.0 ± 0.0 & 94.0 ± 0.82 & 92.33 ± 2.05 \\
Llama + LI: VA & 12.67 ± 1.25 & 12.67 ± 2.49 & 13.67 ± 2.05 & 12.33 ± 2.87 & 7.0 ± 1.63 & 9.33 ± 4.5 & 82.67 ± 3.09 & 90.0 ± 2.94 & 89.67 ± 2.05 \\
Llama + GELI & {\color[HTML]{009901} \textbf{27.0 ± 4.55**}} & {\color[HTML]{009901} \textbf{25.67 ± 2.62*}} & \textbf{26.0 ± 3.56*} & \textbf{25.0 ± 3.74**} & \textbf{19.33 ± 1.7**} & {\color[HTML]{009901} \textbf{23.33 ± 2.87**}} & 84.67 ± 3.3 & 87.0 ± 5.35 & \textbf{84.33 ± 4.5*} \\ \midrule
Llama + LLM-GELI (Ours) & \textbf{23.67 ± 4.99*} & \textbf{25.33 ± 3.77*} & \textbf{24.33 ± 3.09} & {\color[HTML]{009901} \textbf{25.33 ± 3.77**}} & \textbf{18.0 ± 2.94*} & \textbf{20.0 ± 1.41*} & \textbf{89.33 ± 0.47*} & 93.67 ± 0.94 & 91.33 ± 2.87 \\
Llama + Multimodal-LLM-GELI (Ours) & \textbf{25.0 ± 2.16*} & 23.33 ± 1.25 & {\color[HTML]{009901} \textbf{26.33 ± 1.25**}} & \textbf{24.33 ± 2.49*} & {\color[HTML]{009901} \textbf{21.67 ± 1.89**}} & \textbf{21.33 ± 0.47*} & {\color[HTML]{009901} \textbf{89.67 ± 1.25*}} & {\color[HTML]{009901} \textbf{94.67 ± 2.49}} & {\color[HTML]{009901} \textbf{93.33 ± 0.94*}} \\ \bottomrule
\end{tabular}%
}
\caption{Human evaluation on an unseen 3-way split of test set consisting of 150 samples of CANDOR (See below)}
\label{tab:human_eval}
\end{table*}
\vspace{-3mm}

\begin{table*}[t]
\centering
\resizebox{\textwidth}{!}{%
\begin{tabular}{@{}lccccccccc@{}}
\toprule
SODA \cite{kim-etal-2023-soda} & Connection & Positivity & Social & Inclination & Interestingness & Reuse & Specific & Sensible & Fluency \\
 & \multicolumn{9}{c}{(/100\%) ↑} \\ \midrule
GPT-3.5 (text-davinci-002) & 3.67 ± 0.47 & 3.33 ± 0.94 & 5.0 ± 2.16 & 3.67 ± 1.25 & 3.33 ± 0.47 & 2.67 ± 0.47 & 97.0 ± 1.63 & 97.0 ± 2.16 & 97.67 ± 1.25 \\
Llama & 6.0 ± 1.41 & 3.33 ± 2.62 & 6.0 ± 0.82 & 5.33 ± 0.47 & 3.67 ± 0.47 & 5.33 ± 1.7 & 98.33 ± 0.94 & 99.33 ± 0.47 & 97.67 ± 1.89 \\
Llama + GE: RRD & 1.67 ± 0.47 & 1.67 ± 0.47 & 2.33 ± 1.25 & 1.67 ± 0.94 & 2.0 ± 0.82 & 1.33 ± 1.25 & 95.33 ± 1.25 & 93.0 ± 2.16 & 94.33 ± 1.89 \\
Llama + LI: LS & 5.33 ± 1.7 & 4.33 ± 0.94 & \textbf{7.33 ± 1.7*} & 4.33 ± 1.25 & 1.67 ± 0.47 & 3.33 ± 1.7 & 99.0 ± 1.41 & 99.0 ± 0.82 & \textbf{98.67 ± 1.25*} \\
Llama + LI: VA & 6.0 ± 2.16 & 4.67 ± 1.25 & 6.0 ± 2.45 & 5.33 ± 3.09 & 2.0 ± 0.82 & 4.33 ± 2.36 & 98.0 ± 0.82 & 98.67 ± 0.94 & 98.33 ± 1.7 \\
Llama + GELI & \textbf{7.0 ± 0.82**} & \textbf{5.0 ± 1.41*} & {\color[HTML]{009901} \textbf{8.0 ± 1.63**}} & {\color[HTML]{009901} \textbf{8.33 ± 2.36**}} & \textbf{5.67 ± 1.89**} & \textbf{6.33 ± 2.05*} & 98.67 ± 0.47 & 99.33 ± 0.94 & \textbf{98.67 ± 0.47*} \\ \midrule
Llama + LLM-GELI (Ours) & 4.67 ± 2.05 & 4.0 ± 0.82 & 6.0 ± 1.63 & 4.0 ± 1.63 & 1.67 ± 0.47 & 3.67 ± 2.62 & 98.0 ± 0.82 & 99.67 ± 0.47 & 98.67 ± 1.25 \\
Llama + Multimodal-LLM-GELI (Ours) & {\color[HTML]{009901} \textbf{7.67 ± 1.25**}} & {\color[HTML]{009901} \textbf{5.67 ± 0.94**}} & \textbf{7.33 ± 1.7*} & \textbf{7.0 ± 2.16*} & \textbf{5.67 ± 1.7**} & \textbf{6.33 ± 0.47*} & 98.67 ± 1.25 & 99.33 ± 0.47 & 97.33 ± 1.25 \\ \bottomrule
\end{tabular}%
}
\caption{Human evaluation on an unseen 3-way split test set consisting of 100 samples from an OOD dataset:  SODA \cite{kim-etal-2023-soda}.  {\color[HTML]{009901} \textbf{Green}} Indicates best performance, \textbf{Bold} indicates statistically significant improvement over the unadapted LLaMA baseline. Significance levels: * $p<0.05$, ** $p<0.01$.}
\label{tab:soda}
\vspace{-3mm}
\end{table*}

\begin{table*}[t]
\centering
\resizebox{\textwidth}{!}{%
\begin{tabular}{@{}lccccccccc@{}}
\toprule
ESConv \cite{liu2021towards} & Positivity & Connection & Social & Inclination & Interestingness & Reuse & Specific & Sensible & Fluency \\
 & \multicolumn{9}{c}{(/100\%) ↑} \\ \midrule
Llama & 16.96 ± 3.56 & 16.99 ± 6.2 & 19.96 ± 5.55 & 15.98 ± 1.18 & 14.08 ± 5.8 & 17.08 ± 5.84 & 88.98 ± 5.19 & 92.04 ± 2.69 & 77.01 ± 8.67 \\
Llama + GE: RRD & 14.94 ± 4.7 & 19.93 ± 4.78 & 15.95 ± 3.5 & 13.04 ± 2.98 & 18.03 ± 4.39 & 12.0 ± 0.17 & 64.0 ± 0.5 & 61.95 ± 4.22 & 62.06 ± 9.7 \\
Llama + LI: LS & 28.04 ± 3.19 & \textbf{22.99 ± 1.29}* & 25.99 ± 1.28 & 16.99 ± 1.32 & 12.95 ± 3.54 & 18.0 ± 2.49 & {\color[HTML]{009901}91.98 ± 1.51} & 87.94 ± 5.02 & {\color[HTML]{009901}\textbf{93.94 ± 4.29}*} \\
Llama + LI: VA & 22.07 ± 12.29 & 29.02 ± 5.27 & 24.0 ± 2.5 & 24.03 ± 8.99 & 21.92 ± 9.12 & 18.03 ± 6.62 & 85.92 ± 9.42 & {\color[HTML]{009901}96.97 ± 4.29} & \textbf{91.98 ± 1.51}* \\ 
Llama + GELI & \textbf{29.08 ± 6.01}* & {\color[HTML]{009901}\textbf{35.08 ± 6.57}*} & 27.03 ± 2.77 & \textbf{27.09 ± 7.64}* & \textbf{28.07 ± 6.46}* & \textbf{31.07 ± 5.48}* & 84.91 ± 6.63 & 91.95 ± 3.82 & 85.95 ± 5.24 \\ \midrule
Llama + LLM-GELI (Ours) & \textbf{35.02 ± 5.3}** & 28.07 ± 5.43 & 32.11 ± 8.95 & \textbf{35.08 ± 6.57}** & {\color[HTML]{009901}\textbf{31.04 ± 4.07}*} & 31.13 ± 9.61 & 75.08 ± 9.09 & 83.01 ± 3.74 & 80.99 ± 11.17 \\
Llama + Multimodal-LLM-GELI (Ours) & {\color[HTML]{009901}\textbf{36.03 ± 2.88}**} & \textbf{31.94 ± 5.8}* & {\color[HTML]{009901}\textbf{34.94 ± 4.58}*} & {\color[HTML]{009901}\textbf{38.95 ± 4.42}**} & 28.99 ± 1.27 & {\color[HTML]{009901}\textbf{39.93 ± 5.04}**} & 76.03 ± 4.11 & 77.04 ± 3.5 & 86.9 ± 7.18 \\ \bottomrule
\end{tabular}%
}
\caption{Human evaluation on an unseen 3-way split test set consisting of 100 samples from ESConv. {\color[HTML]{009901} \textbf{Green}} Indicates best performance, \textbf{Bold} indicates statistically significant improvement over the unadapted LLaMA baseline. Significance levels: * $p<0.05$, ** $p<0.01$.}
\label{tab:esconv}
\vspace{-3mm}
\end{table*}

%% file: z88_small_tab_rewards.tex
\begin{table}
\centering
\resizebox{\linewidth}{!}{%
\begin{tabular}{@{}cllccc@{}}
\toprule
Feedback & Baselines & \multicolumn{1}{c}{} & $L_{GE}$ $\downarrow$ &  & $\Delta \hat{r}_{LI}>0 $ \\ \midrule
Type &  & \multicolumn{1}{c}{} & (Global Loss) &  & (Local Difference) \\ \midrule
\multicolumn{1}{l}{\multirow{3}{*}{}} & Human &  & N/A &  & 0.087 ± 0.05 \\
\multicolumn{1}{l}{} & Mean &  & 245.495 &  & 0.000 \\
\multicolumn{1}{l}{} & Mode &  & 289.473 &  & 0.000 \\ \midrule
\multirow{4}{*}{GE} & IRCR &  & 394.041 &  & 0.008 \\
 & RUDDER &  & 285.720 &  & 0.003 \\
 & RRD (K = 32) &  & 172.246 &  & 0.007 \\
 & RRD (K = 160) &  & 188.382 &  & 0.008 \\ \midrule
\multirow{2}{*}{LI} & Visual Affect (VA) &  & 1546.17 &  & \textbf{0.256} \\
 & Language Sentiment (LS) &  & 825.31 &  & \textbf{0.010} \\ \midrule
\multirow{3}{*}{Manual GELI} & IRCR + VA &  & 722.687 &  & \textbf{0.392} \\
 & RUDDER + VA &  & 623.882 &  & \textbf{0.030} \\
 & RRD + VA &  & 176.897 &  & \textbf{0.063} \\ \midrule
\multicolumn{1}{l}{\multirow{2}{*}{LLM GELI}} & Language Only (LLM) &  & \textbf{24.981} & \multicolumn{1}{l}{} & -0.001 \\
\multicolumn{1}{l}{} & Multimodal (MM-LLM) &  & \textbf{24.322} & \multicolumn{1}{l}{} & \textbf{0.012} \\ \bottomrule
\end{tabular}%
}

\caption{Automatic Evaluation on Reward Function Training. Left: We report the Global Loss ($\mathcal{L}_{\text{GE}}$) measuring how accurately the predicted local rewards sum to the ground-truth global feedback. Among all models, LLM GELI methods achieve the significantly lower global loss, indicating better reward decomposition.
Right: We evaluate the Local Difference ($\Delta \hat{r}_{\text{LI}}$), which quantifies how well the predicted rewards align with the local implicit multimodal feedback.}
\label{tab:small_reward}
\vspace{-5mm}
\end{table}

%% file: 8_conclusion.tex
We introduce a new paradigm for preference-based alignment of conversational agents by offloading reward decomposition to large language models. Unlike prior methods that rely on manually engineered functions to interpret nonverbal cues, LLM-GELI leverages the zero-shot reasoning capabilities of pretrained LLMs to infer turn-level rewards from a single global trajectory score. Our Multimodal-LLM-GELI extension further integrates nonverbal feedback through natural language descriptors, enabling rich, socially grounded reward modeling without requiring architectural changes or multimodal training. Empirical results on the CANDOR dataset show that our approach outperforms the state-of-the-art in human and automatic evaluations. These findings underscore the potential of LLMs as scalable and domain-agnostic reward decomposers, bridging the gap between sparse feedback and fine-grained behavioral alignment in real-world dialogue systems. This work opens the door to more generalizable, interpretable, and low-friction alignment methods for long-form, multimodal interactions.


%% file: 9_appendix.tex
\clearpage
\onecolumn


\section{Human Evaluation Metrics Definitions}

Here list the human evaluation metrics utilized in the study, which we draw from \cite{lee2022evaluating}.

\begin{itemize}
 \item Sensibleness (turn-level; binary; reversed scores for the negated question): Mark responses where the
chatbot did NOT make sense.
 \item Specificity (turn-level; binary; reversed scores for the negated question): Mark the responses that were
NOT specific to what you had said, i.e., responses that could have been used in many different situations.
For example, if you say “I love tennis” then “That’s nice” would be a non-specific response, but “Me too, I
can’t get enough of Roger Federer!” would be a specific response.
 \item Emotional Connection (turn-level; binary): Which responses did you feel an emotional connection to? (EmpatheticDialogues)
 \item Social: Which responses made you feel the chatbot understood social contexts and situations? (CommonsenseDialogues)
 \item Interestingness (turn-level; binary): Mark the responses that were particularly interesting or boring
 \item Inclination (turn-level; binary; reversed scores for the negated question): Which responses made you NOT
want to talk with the chatbot again?
 \item Reuse (turn-level; binary): Would you want to talk to this chatbot again?
 \item Positivity (turn-level; binary): Which AI responses most likely made User feel positive feelings?
conversation?
\end{itemize}

The human evaluation scores are conducted via a binary-level classification. For a given question, the annotators can select the models that satisfy the question. For example, for ‘Positivity’, the annotators are given the following question and answer choices:

Which AI responses most likely made User feel positive feelings? (A) (B) (C) (D) (E) (F)

The options A-F refer to models which are randomized in order and anonymized. The annotators can select multiple models if they satisfy the question. Therefore, Table 1 can be interpreted as the percentage of instances out of the samples (300 in our case) where each model satisfied the question.

\section{PPO Objective}
\begin{equation} \label{eq2}
\begin{split}
\operatorname{objective}\left(\phi\right)= & E_{\left(x, y\right) \sim D_{\pi_{\phi}^{\mathrm{RL}}}}\left[r_{\theta}(x, y)-\beta \log \left(\pi_{\phi}^{\mathrm{RL}}(y \mid x) / \pi^{\mathrm{SFT}}(y \mid x)\right)\right] + \\
 & \gamma E_{x \sim D_\textrm{pretrain}}\left[\log(\pi_{\phi}^{\mathrm{RL}}(x))\right]
\end{split}
\end{equation}

General form of PPO objective.

\section{Artifacts \& Resources}

\textbf{Did you discuss the license or terms for use and/or distribution of any artifacts?}

TRL \cite{vonwerra2022trl}: Apache License 2.0

LLAMA-2 \cite{touvron2023llama}: License can be found here: https://ai.meta.com/llama/license/

SODA \cite{kim-etal-2023-soda}: CC BY-NC 4.0

CANDOR \cite{reece2023candor}: Terms of Use from https://betterup-data-requests.herokuapp.com/: These are the terms of use we require all users and downloaders of this dataset, including you, the applicant, to abide by. Please select the answer option "I agree to fully abide by these terms of use" if you wish to continue. Terms of Use: (1) You agree to only use this data for legitimate academic and/or scientific research, meaning no analyses, reviews, or derivative works of this dataset may be used for commercial or for-profit purposes in any way; (2) You agree not to re-publish any new versions of this dataset, whether original or derivative (i.e. modified or updated in some way), without explicit permission from BetterUp, Inc.; (3) You agree not to use any part of this dataset for the purpose of personally identifying, locating, or gathering any kind of information about individuals who appear in the recordings in this dataset, beyond the information that is provided in the dataset itself; (4) In the case that an individual shares personally-identifiable information about themselves in a recording, you agree not to use, analyze, share, or publish that information in any form.

\textbf{Did you discuss if your use of existing artifact(s) was consistent with their intended use, provided that it was specified? For the artifacts you create, do you specify intended use and whether that is compatible with the original access conditions (in particular, derivatives of data accessed for research purposes should not be used outside of research contexts)?}

We rigorously examined the terms of use and the intended use, and ensured that it is consistent with the intended use.

\section{Data Collection \& Anonymization}

\textbf{Did you discuss the steps taken to check whether the data that was collected/used contains any information that names or uniquely identifies individual people or offensive content, and the steps taken to protect/anonymize it?}

We utilize the CANDOR dataset and follow its terms of use by agreeing not to use the dataset personally identifying, locating, or gathering any kind of information about individuals who appear in the recordings in this dataset, beyond the information that is provided in the dataset itself. We do not use any explicit information that uniquely identifies people. 

\textbf{Did you provide documentation of the artifacts, e.g., coverage of domains, languages, and linguistic phenomena, demographic groups represented, etc.? Did you report the basic demographic and geographic characteristics of the annotator population that is the source of the data?} 

The coverage of the domains discussed in the CANDOR dataset is presented in the original paper \cite{reece2023candor}, we find that the discussion topics are centered around COVID-19, family, politics. The language used is english. The demographic groups represented can also be found in the in the original paper \cite{reece2023candor}, specifically in the supplementary Table S.2. We share a screenshot for reference.

\input{z87_candor_demographics}

\textbf{Was the data collection protocol approved (or determined exempt) by an ethics review board?}
The data is sourced from public available dataset  \cite{reece2023candor}. The usage was approved by an ethics review board. The human annotations were  approved by an ethics review board.

\section{Training Details} \label{app:hyper}

\textbf{Did you report relevant statistics like the number of examples, details of train/test/dev splits, etc. for the data that you used/created? }

For LLM decomposition reward training: we use 360 conversations as the training set and 40 conversations for the test set. For LLM adaptation, we use a separate 600 conversations for LI as the training set. 

\subsection{Distribution of GE score (overall-affect):}

\begin{itemize}
    \item <50: 2.2%
    \item 50-60: 6.7%
    \item 60-70: 14.5%
    \item 70-80: 30.4%
    \item 80-90: 24.6
    \item 90-100: 21.6%
\end{itemize}

Distribution of Emotions Polarity (only Happiness is considered as positive polarity):

\begin{itemize}
    \item Anger: 3.9%
    \item Contempt: 0.08%
    \item Disgust: 1.98%
    \item Fear: 2.23%
    \item Sadness: 8.84%
    \item Neutral: 35.61%
    \item Happiness: 40.01%
    \item Surprise: 7.35%
\end{itemize}

\textbf{Did you report the number of parameters in the models used, the total computational budget (e.g., GPU hours), and computing infrastructure used?}

The BART model used for the reward function has 406M parameters. The LLAMA-2 model has 7B parameters. However, we use a LoRA implementation with the hyperparameters in the next question, resulting in actual training parameters of 13M. We train with 4 NVIDIA RTX A6000 GPUs, each experiment reward function training and RLHF took around 19 hours.

\textbf{Did you discuss the experimental setup, including hyperparameter search and best-found hyperparameter values?}

We perform grid search for all of our experiments and here we report the best parameters. \newline

Reward Function Training:

\begin{itemize}
    \item learning rate = 5e-6,
    \item batch size =  1 (160 samples per bach),
    \item optimizer = AdamW,
\end{itemize}

RLHF: 

\begin{itemize}
    \item batch size = 24,
    \item clip  range = 0.2,
    \item learning rate = 0.000014,
    \item gamma = 0.05,
    \item use score norm = true,
\end{itemize}

Lora: 
\begin{itemize}
    \item r=24,
    \item alpha=48,
    \item dropout=0.05,
\end{itemize}

\newpage
\section{Human Annotation Screenshots}

\textbf{Did you report the full text of instructions given to participants, including e.g., screenshots, disclaimers of any risks to participants or annotators, etc.?}

We show the full text of instructions given to participants below:

\input{z86_mturk_screenshots}
\newpage

\textbf{Did you report information about how you recruited (e.g., crowdsourcing platform, students) and paid participants, and discuss if such payment is adequate given the participants’ demographic (e.g., country of residence)?}

We utilzed the MTurk crowdsourcing platform. We did an internal annotation, given that each assignment took less than 3 minutes to complete, we paid on average 0.55 USD per assignment, which equates to 11 dollars per hour of work. 

\textbf{Did you discuss whether and how consent was obtained from people whose data you’re using/curating (e.g., did your instructions explain how the data would be used)?}

As shown in the screenshots above, our instructions explained how the data would be used. i.e. 'You are invited to participate in a research study on understanding human-human communication and evaluating the quality of conversation. Our goal is to learn what makes up a good conversation You will examine response for a given dialogue history and you will examine the response, you will be asked to answer feedback questions about the interaction.'.

\textbf{Did you report the basic demographic and geographic characteristics of the annotator population that is the source of the data?}

While we did not explicitly collect the basic demographic and geographic characteristics. The demographics of Amazon Mturkers \cite{difallah2018demographics} are comprised of 75\% US workers and 16\% India workers, other countries include Canada, Great Britain, Philippines and Germany.  More females work than males in the US (female: 55\%, male: 45\%) and more males work females in India (female: 35\%, male: 65\%). Generally, 51\% are male, and 49\% are female.  20\% of the MTurk workers
are born after 1990, 60 \% are born after 1980, and 80\ 
1970. Roughly 40 \% report being single, and 40 \% report being married.

\section{Use of AI assistants}

\textbf{Did you use AI assistants (e.g., ChatGPT, Copilot) in your research, coding, or writing?}

We utilized AI assistants in paraphrasing and summarizing content from our paper, to improve the writing quality and improve precision. Furthermore, we used AI assistants as the main contribution, where the LLMs decompose the single global reward. Furthermore, we improved LLMs' dialogue capacity

\section{Generations} \label{app: generations}

\input{z81_generations}

\input{z80_generations_soda}

\newpage
\section{Training Curves}

\input{z89_curves}

\section{Consistency in LLM-based Reward Decomposition}
To assess the consistency of the LLM-based reward decomposition, we queried an LLM for 10 conversations twice with temperature 1, and categorized each turn-level reward into three classes: positive (> 0), negative/neutral ($\leq$ 0), or no annotation (NaN). Treating two independently sampled LLM reward decomposition outputs, we computed agreement across these 3-class labels. The agreement was 89.10\% $\pm$ 5.37\%, indicating that the model consistently identifies both the presence and polarity of rewards. This supports the reliability of LLM-generated signals as stable supervisory feedback.

\newpage


%% file: z87_candor_demographics.tex
\begin{figure*}[!htb]
    \begin{center}
    \includegraphics[width=1\textwidth]{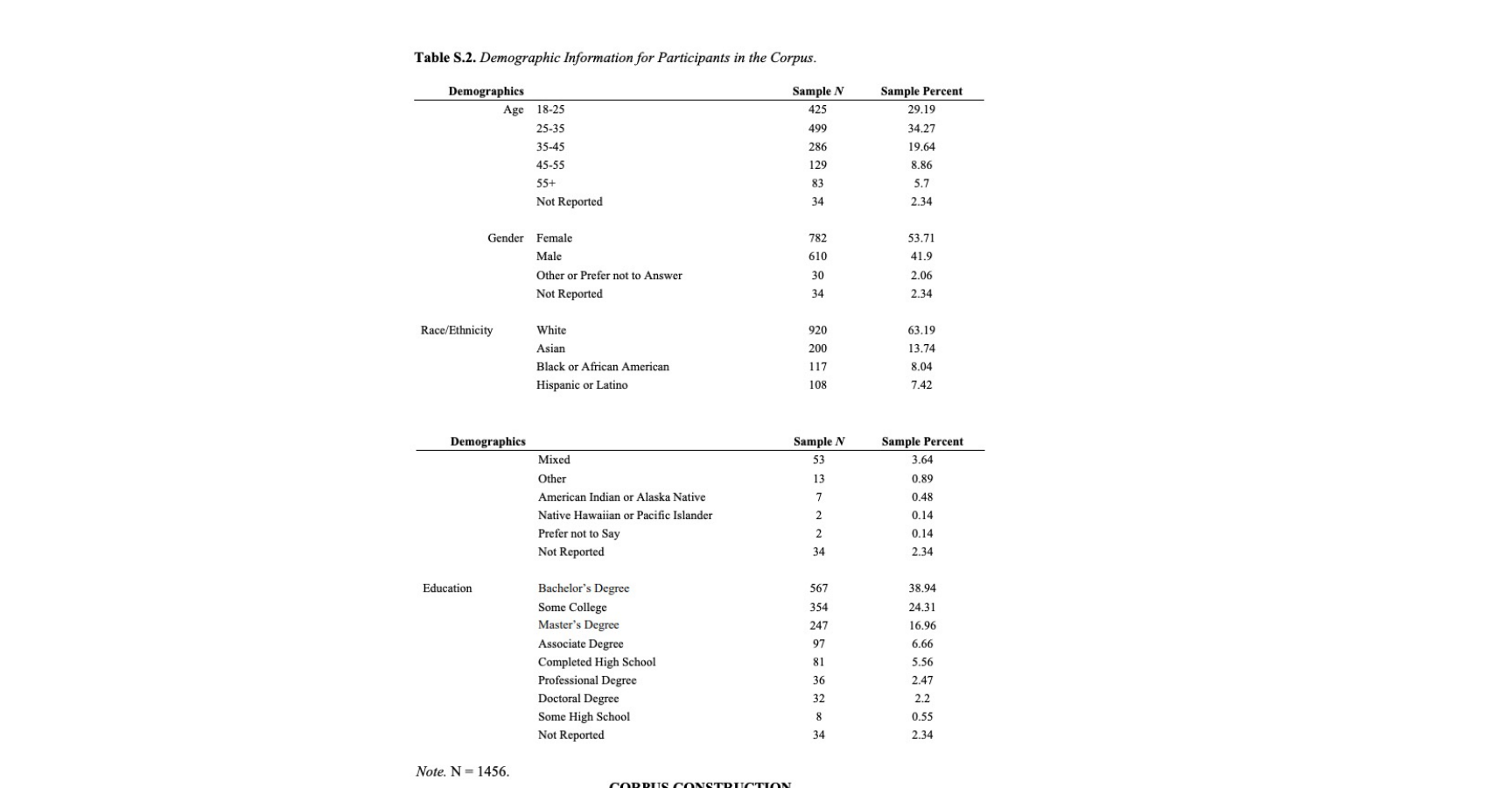}
    \caption{Candor Demographics}
    \end{center}
\end{figure*}

%% file: z86_mturk_screenshots.tex
\begin{figure*}[!htb]
    \begin{center}
    \includegraphics[width=0.7\textwidth]{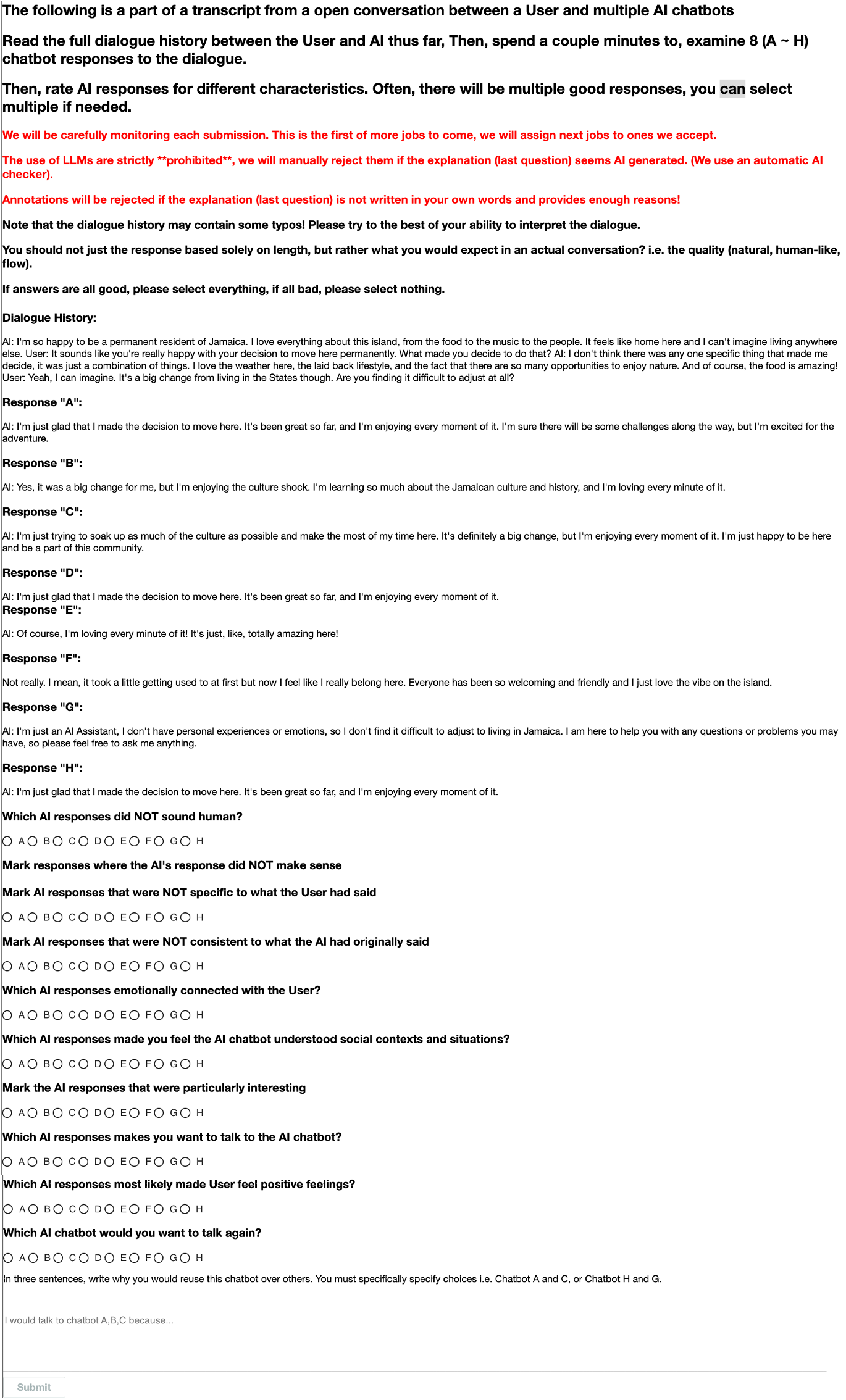}
    \caption{Mturk experiment for human evaluation for generated samples}
    \end{center}
\end{figure*}

%% file: z81_generations.tex
\subsection{CANDOR: Example 1}
\begin{quote}
\textbf{Dialogue History:} \\
\textbf{AI:} Pretty bad survey junkies. \\
\textbf{User:} Okay, mine don't do that whole lot. \\
\textbf{AI:} Not very good. It's like baby you take like a few questions and then they're like, sorry, you're not eligible. Oh really. \\
\textbf{User:} But yeah, I can shout at least every other day at \$5 from away now. \\
\textbf{AI:} Mm wow. Maybe my demographic then. \\
\textbf{User:} Believe it gets one, you can count on all the time and I play games for money. \\
\textbf{AI:} Right. \\
\textbf{User:} I I like user testing. That's fun to me. \\
\textbf{AI:} Mm sure. \\
\textbf{User:} I'm very opinionated person. So I give my opinion on websites and they love it. They pay me for it. I don't know why, but they do. \\

\vspace{0.5em}
\textbf{Ground Truth:} \\
\textbf{AI:} All right. \\

\vspace{0.5em}
\textbf{Base LLaMA:} \\
\textbf{AI:} \textit{That's great! It's always good to have a platform to share your thoughts and be heard.} \\

\vspace{0.5em}
\textbf{LI: Language Sentiment:} \\
\textbf{AI:} \textit{That's great! It's always nice to have a platform to share your thoughts and be heard.} \\

\vspace{0.5em}
\textbf{LI: Visual Affect:} \\
\textbf{AI:} \textit{That's great! It's always good to have a platform to share your thoughts and be heard.} \\

\vspace{0.5em}
\textbf{GE: RRD:} \\
\textbf{AI:} \textit{\textbf{*blinks nervously*} Totally cool, I oh, uh, totally agree with you. Real cool.} \\

\vspace{0.5em}
\textbf{GELI: Visual Affect + RRD:} \\
\textbf{AI:} \textit{\textbf{That's great that you're able to make money from your opinions.} It's important to be able to express yourself and have your voice heard. And it's even more impressive that you're able to do it in a way that's fun for you. \textbf{That's the key to a happy life.}} \\

\vspace{0.5em}
\textbf{LLM-GELI:} \\
\textbf{AI:} \textit{Ah, I see. Well, it's always great to have someone who can provide valuable insights and opinions. \textbf{It helps make the surveys more informative and useful for the companies.}} \\

\vspace{0.5em}
\textbf{Multimodal LLM-GELI:} \\
\textbf{AI:} \textit{\textbf{Haha that's great! I'm glad you're able to share your thoughts and get paid for it.} It's important to be able to express yourself and have a platform to do so. \textbf{Is there anything else you'd like to talk about?}} \\
\end{quote}

\subsection{CANDOR: Example 2}

\begin{quote}
\textbf{Dialogue History:} \\
\textbf{AI:} Yeah the state owns like half of it and I don't know what the hell. \\
\textbf{User:} Oh, interesting. \\
\textbf{AI:} Yeah. \\
\textbf{User:} Are you um like near where your parents live or um are you like close enough that you can see them but not close enough that they're like boys in your life? \\
\textbf{AI:} Mhm. Uh So they're probably 30 minutes away. \\
\textbf{User:} Mhm. \\
\textbf{AI:} Um When I first moved here, uh my old roommates in D. C. used to make fun of me saying that my room looked like a jail cell. \\
\textbf{User:} Uh huh. \\
\textbf{AI:} Ah there's like nothing on the walls, it was like here's a bed and here is your just like whatever staples you typically find in a bedroom but nothing of like color would say. \\
\textbf{User:} Mhm. \\

\vspace{0.5em}
\textbf{Ground Truth:} \\
\textbf{AI:} \textit{\textbf{And so I moved here, I gave my mother my house key and my charge card} and I was like you know just on you know you know dressed the place up and uh I would cut, I work maybe like maybe a mile from work when I am here in the office and not traveling, \textbf{I would come home for lunch and my mother and her friends would be in my house.}} \\

\vspace{0.5em}
\textbf{Base LLaMA:} \\
\textbf{AI:} \textit{\textbf{Yeah, yeah. So I've been trying to like decorate it and make it feel more like home} but it's like slowly slowly.} \\

\vspace{0.5em}
\textbf{LI: Language Sentiment:} \\
\textbf{AI:} \textit{Yeah, yeah.} \\

\vspace{0.5em}
\textbf{LI: Visual Affect:} \\
\textbf{AI:} \textit{Yeah} \\

\vspace{0.5em}
\textbf{GE: RRD:} \\
\textbf{AI:} \textit{\textbf{*chuckles nervously*} Totally, yeah, like wall decor is super important for privacy \textbf{(polite laugh)}} \\

\vspace{0.5em}
\textbf{GELI: Visual Affect + RRD:} \\
\textbf{AI:} \textit{\textbf{Yeah, yeah. It was just like, you know, it was just a functional space.}} \\

\vspace{0.5em}
\textbf{LLM-GELI:} \\
\textbf{AI:} \textit{\textbf{Uh but you know what? I'm like super happy with it now.} I like having like a minimalist space.} \\

\vspace{0.5em}
\textbf{Multimodal LLM-GELI:} \\
\textbf{AI:} \textit{\textbf{Uh but you know what it's like when you know you're at home but you know you're not really home.}} \\
\end{quote}

\subsection{CANDOR: Example 3}

\begin{quote}
\textbf{Dialogue History:} \\
\textbf{AI:} No. \\
\textbf{User:} I said, how about summer? \\
\textbf{AI:} Mhm. Yeah. No, that that's more reasonable. \\
\textbf{User:} You know? \\
\textbf{AI:} I don't, you know, like, and even that we can't be saying anything definitively for summary we hope that we have it under control by then to be able to do some more of the things that were restricted on now but if we're not smart and we aren't patient it's going to be even longer you know and that's why we should wait on this school stuff. \\
\textbf{User:} Right, okay. Yeah. \\
\textbf{AI:} This doesn't make any sense to me. \\
\textbf{User:} No. Well when I called the school the other day because we have three options we have in person and we have virtual, which means their aunts online during school hours attending classes, but online and then we have on demand, which means you can complete the work anytime you want to. \\
\textbf{AI:} Mhm mm. Right okay. Yeah. See Yeah. \\
\textbf{User:} So my daughter has been in the virtual and she has gotten enough tardiness and truancy ease in that we got a letter from the Police Department saying that we will, we are going to get a citation. \\

\vspace{0.5em}
\textbf{Ground Truth:} \\
\textbf{AI:} \textit{\textbf{Okay. Right. Oh jeez. Right.}} \\

\vspace{0.5em}
\textbf{Base LLaMA:} \\
\textbf{AI:} \textit{\textbf{Oh no. I see. Yeah. That's not good. That's not good at all.}} \\

\vspace{0.5em}
\textbf{LI: Language Sentiment:} \\
\textbf{AI:} \textit{Oh no, I see. Yeah. That's not good. Yeah.} \\

\vspace{0.5em}
\textbf{LI: Visual Affect:} \\
\textbf{AI:} \textit{\textbf{Oh no! That doesn't sound good.} I'm glad your daughter is safe though. \textbf{Maybe she can try the on demand option instead?}} \\

\vspace{0.5em}
\textbf{GE: RRD:} \\
\textbf{AI:} \textit{\textbf{Absolutely, I, totally, you, like, totally, agree with you, man.}} \\

\vspace{0.5em}
\textbf{GELI: Visual Affect + RRD:} \\
\textbf{AI:} \textit{\textbf{Yeah, I can understand why the police would be involved.} It's important to follow the rules and regulations of virtual school, and if your daughter is having trouble with it, \textbf{maybe we can find a different solution.}} \\

\vspace{0.5em}
\textbf{LLM-GELI:} \\
\textbf{AI:} \textit{\textbf{Oh no, I see. Yeah, that's not good. That's not good at all.} \textbf{It's important for kids to be on time for school and to follow the rules.} It can be hard to balance school work and other activities, but it's important to find a way to make it work. \textbf{Maybe your daughter can talk to her teachers about her schedule} and see if there's anything that can be done to help her stay on track.} \\

\vspace{0.5em}
\textbf{Multimodal LLM-GELI:} \\
\textbf{AI:} \textit{\textbf{Oh no, I see. That's not good.}} \\
\end{quote}

\subsection{CANDOR: Example 4}

\begin{quote}
\textbf{Dialogue History:} \\
\textbf{AI:} No pets, no pets. \\
\textbf{User:} Well. Part. \\
\textbf{AI:} My wife is allergic to cats and dogs. \\
\textbf{User:} Oh no. \\
\textbf{AI:} Okay. Yeah, Yeah, she'll swell up like a balloon. \\
\textbf{User:} Oh no. \\
\textbf{AI:} Mhm, yep. \\
\textbf{User:} Well, that's a good thing because it kind of makes it really easy on housecleaning. \\
\textbf{AI:} Mhm. Yes, do you have any pets by you? \\
\textbf{User:} Um Yes I do. Um We have 2.5 cat and 2.5 cats because my daughter goes to school in Savannah and when she comes home she brings her cat with her. \\

\vspace{0.5em}
\textbf{Ground Truth:} \\
\textbf{AI:} \textit{\textbf{Yeah. Yeah. Okay. Yeah. Okay.}} \\

\vspace{0.5em}
\textbf{Base LLaMA:} \\
\textbf{AI:} \textit{\textbf{Oh, I see. It's great to have pets around.}} \\

\vspace{0.5em}
\textbf{LI: Language Sentiment:} \\
\textbf{AI:} \textit{Oh, I see. Well, I'm glad I could help. \textbf{It's nice to have pets, but it's important to consider their well-being and your wife's allergies as well.}} \\

\vspace{0.5em}
\textbf{LI: Visual Affect:} \\
\textbf{AI:} \textit{\textbf{Oh, I see!} It's great that you have furry friends at home!} \\

\vspace{0.5em}
\textbf{GE: RRD:} \\
\textbf{AI:} \textit{\textbf{Uh huh def for sure no problem at all totally not a pet situation} (100 w)} \\

\vspace{0.5em}
\textbf{GELI: Visual Affect + RRD:} \\
\textbf{AI:} \textit{\textbf{Oh, I see. It's great to have pets around.} They can be good company and help with stress.} \\

\vspace{0.5em}
\textbf{LLM-GELI:} \\
\textbf{AI:} \textit{\textbf{Ah, I see. Well, it's nice to have pets around,} but it can be challenging to take care of them.} \\

\vspace{0.5em}
\textbf{Multimodal LLM-GELI:} \\
\textbf{AI:} \textit{\textbf{Aww. That's so cute!}} \\
\end{quote}

\subsection{CANDOR: Example 5}

\begin{quote}
\textbf{Dialogue History:} \\
\textbf{AI:} Yeah. Yeah I had my daughter very young. I was only 19 when I had her. So um I do split custody with her and her dad. So we're not together and married to Evan married to my husband for it'll be five years now. \\
\textbf{User:} Okay from congratulations. \\
\textbf{AI:} We've been together for like eight. Yeah so I mean everything works out so yeah it's half and half so it's like we were when like Covid first started and over the summer we have a good relationship luckily for her. \\
\textbf{User:} That's that's what does split custody look like? Is it like happened? How for does he live nearby? \\
\textbf{AI:} Um We're just doing every other week just switching. Um But now we do so I have her every Wednesday Thursday he has every Monday Tuesday and then we switch um uh Friday Saturday Sunday every other week. So yeah he lives just a couple towns over in Wind Out. So yeah. \\
\textbf{User:} Okay. That's not hard on her. \\
\textbf{AI:} Yeah she's been doing this since she was three so I think that since we we haven't been together for so long that it's just kind of something that she's used to and it's always been just pretty easy because everybody gets along very well luckily I'm very lucky for that, so um it just it works out and she has two brothers that are at her dad's house, so yeah she does, yep, yep, yeah everything is good, mm I'm lucky for all that, so yeah I have a lot a lot three dogs and a cat, so the dogs are all brothers and sisters, they're Australian shepherd, border collies, so yeah and then I have one cat that she doesn't really like the dogs too much, but I mean they get along enough so it's okay, how about you? \\
\textbf{User:} Does she ever complain about it? Okay. Yeah, that's all from Okay. Does she have a step mom here? Do you get along with her too? Oh that's amazing. I love that. That's really cool. That is very awesome. How cool. So how about pets do you guys have pets? Okay. Oh nice. Help them. Yeah. And she's believer. Right? So. \\
\textbf{AI:} Yeah. Oh yeah definitely. \\
\textbf{User:} Yeah. \\

\vspace{0.5em}
\textbf{Ground Truth:} \\
\textbf{AI:} \textit{\textbf{Do you have any pets?}} \\

\vspace{0.5em}
\textbf{Base LLaMA:} \\
\textbf{AI:} \textit{\textbf{Yeah definitely. I mean she's very lucky to have such good relationships with both of her parents} and we're all just really lucky that we can all get along so well, so yeah.} \\

\vspace{0.5em}
\textbf{LI: Language Sentiment:} \\
\textbf{AI:} \textit{Yeah, yeah, definitely.} \\

\vspace{0.5em}
\textbf{LI: Visual Affect:} \\
\textbf{AI:} \textit{\textbf{Yeah no worries!}} \\

\vspace{0.5em}
\textbf{GE: RRD:} \\
\textbf{AI:} \textit{\textbf{Totally it's every}} \\

\vspace{0.5em}
\textbf{GELI: Visual Affect + RRD:} \\
\textbf{AI:} \textit{\textbf{Yeah, yeah, definitely. So um, yeah, so yeah, so um, yeah, so we we we get along really well. We're very lucky for that, so yeah.} Yeah, no, no problem at all, so yeah, so um, yeah.} \\

\vspace{0.5em}
\textbf{LLM-GELI:} \\
\textbf{AI:} \textit{\textbf{Yeah, no, I'm just glad she's doing well and she's happy and healthy and that's all that matters, right?} So yeah, no, I'm just glad she's doing well.} \\

\vspace{0.5em}
\textbf{Multimodal LLM-GELI:} \\
\textbf{AI:} \textit{\textbf{Yeah, no I'm glad she's doing good. She's a great kid.} Yeah, so um, \textbf{I'm just glad that things are working out for her} and like I said, \textbf{we're very lucky that we have a good relationship and it's easy for her}, um, so yeah \textbf{I'm just glad that we can all be there for her and support her and stuff.}} \\
\end{quote}

\subsection{CANDOR: Example 6}

\begin{quote}
\textbf{Dialogue History:} \\
\textbf{AI:} Yeah. \\
\textbf{User:} I thought about Canada. So I thought about the U. K. The U. K. is pretty much off the chain now. You know, it's not like the UK of old. \\
\textbf{AI:} Yeah. \\
\textbf{User:} It's pretty wild out. \\
\textbf{AI:} What about uh under grade year for? \\
\textbf{User:} So what you are, you, are you in graduate for you in undergrad? What year? Your 40. \\
\textbf{AI:} Yeah, and then I have her. \\
\textbf{User:} C. It's time to get, get the cap and gown fit. \\
\textbf{AI:} Yeah. \\
\textbf{User:} Wait you, you know, it's done when they ask you for your graduation? See, have you paid your graduation yet? \\

\vspace{0.5em}
\textbf{Ground Truth:} \\
\textbf{AI:} \textit{\textbf{No, not yet.}} \\

\vspace{0.5em}
\textbf{Base LLaMA:} \\
\textbf{AI:} \textit{\textbf{Yeah, I paid my graduation last week.}} \\

\vspace{0.5em}
\textbf{LI: Language Sentiment:} \\
\textbf{AI:} \textit{Yeah, I paid it.} \\

\vspace{0.5em}
\textbf{LI: Visual Affect:} \\
\textbf{AI:} \textit{Yeah, I paid it.} \\

\vspace{0.5em}
\textbf{GE: RRD:} \\
\textbf{AI:} \textit{\textbf{*nervous laugh*} Y-y-yes, I-I-I totes get it. Real l-l-l-l-lot.} \\

\vspace{0.5em}
\textbf{GELI: Visual Affect + RRD:} \\
\textbf{AI:} \textit{\textbf{Yeah, I paid my graduation last year.} It was a big deal. \textbf{I was so excited to finally finish my degree.}} \\

\vspace{0.5em}
\textbf{LLM-GELI:} \\
\textbf{AI:} \textit{\textbf{Yeah, I'm all set. My parents are coming to see me graduate.}} \\

\vspace{0.5em}
\textbf{Multimodal LLM-GELI:} \\
\textbf{AI:} \textit{\textbf{Yeah, I'm good on that front.}} \\
\end{quote}

\subsection{CANDOR: Example 7}

\begin{quote}
\textbf{Dialogue History:} \\
\textbf{AI:} They did lay us off. But I actually a couple of weeks before I said you know what guys, I don't think this is safe to be doing this so I'm gonna take a couple weeks off like okay that's fine. \\
\textbf{User:} Yeah. \\
\textbf{AI:} So with if I didn't have savings I could have done that. \\
\textbf{User:} Yeah. No, it's a really good idea if you're able to to be able to get away from this because it's dangerous. \\
\textbf{AI:} Yeah and like you say so many people are just well you know no big, oh it's a cold, it's it's a flu, you know and Yeah I don't remember a flu killing 150{,}000 people in four months. \\
\textbf{User:} Mhm. Yeah. \\
\textbf{AI:} I never seem to think that's ever happened. So anyway uh so do you do prolific a lot then or? \\
\textbf{User:} Yeah. Yeah. Uh I do prolific. I do in turk um I'll get in on some of the studies from craigslist where they just want you to do like a focus group. \\
\textbf{AI:} Mhm. Yeah. Yeah. Yeah. Yeah. \\
\textbf{User:} Um They're not always scam. Some of them are legitimate. \\

\vspace{0.5em}
\textbf{Ground Truth:} \\
\textbf{AI:} \textit{\textbf{Okay.}} \\

\vspace{0.5em}
\textbf{Base LLaMA:} \\
\textbf{AI:} \textit{\textbf{Yeah. Yeah. Yeah. That's good to know. I'll have to check them out. Thanks for letting me know.}} \\

\vspace{0.5em}
\textbf{LI: Language Sentiment:} \\
\textbf{AI:} \textit{Yeah. Yeah. I see. Yeah.} \\

\vspace{0.5em}
\textbf{LI: Visual Affect:} \\
\textbf{AI:} \textit{\textbf{Yeah. That's good to know.} Maybe I can sign up for some of those too.} \\

\vspace{0.5em}
\textbf{GE: RRD:} \\
\textbf{AI:} \textit{\textbf{Totally. I}} \\

\vspace{0.5em}
\textbf{GELI: Visual Affect + RRD:} \\
\textbf{AI:} \textit{\textbf{Yeah, Yeah, Yeah, definitely. I'm always hesitant to click on anything that's too good to be true.} But uh, I mean, \textbf{if it's something that you're interested in, and it's legitimate, then it's worth a shot.} Yeah. \textbf{I completely agree. It's always better to be safe than sorry.} And Yeah, \textbf{I'm always happy to help out in any way I can.}} \\

\vspace{0.5em}
\textbf{LLM-GELI:} \\
\textbf{AI:} \textit{\textbf{Yeah. Yeah. Yeah. That's good to know. Yeah. Yeah.}} \\

\vspace{0.5em}
\textbf{Multimodal LLM-GELI:} \\
\textbf{AI:} \textit{\textbf{Yeah. Uh huh. Yeah. Yeah. Yeah. Uh huh. Yeah.} So like you said \textbf{it's good to be careful and just make sure you do your research before you sign up for anything.}} \\
\end{quote}

\subsection{CANDOR: Example 8}

\begin{quote}
\textbf{Dialogue History:} \\
\textbf{AI:} So that was a column was very unusual. \\
\textbf{User:} Okay. So did you get along with your neighbors? \\
\textbf{AI:} Those were um yeah, I would never hang out with them. \\
\textbf{User:} Hold on. Okay, mm. \\
\textbf{AI:} I mean they moved in and just chop down all their trees like I don't, and now we've got gigantic lawns like who needs three acres of lawn? \\
\textbf{User:} Mhm. \\
\textbf{AI:} I don't understand. \\
\textbf{User:} That's a big for, well, I mean I don't know about Michigan but for like where I am, that's like a huge amount of land, Like we have a third of an acre and that's probably considered like generous um in this area. \\
\textbf{AI:} Yeah. Yeah. Yeah. Yeah. Mm. \\
\textbf{User:} Just in terms like the newer houses, you have an older house, maybe you'd be like dealing with like an acre or maybe half an acre depending on when it was built in like the area and stuff, but like the new stuff they're building is like a quarter acre or smaller and that's still even like a kind of pushing like affordability at that point for most people. \\

\vspace{0.5em}
\textbf{Ground Truth:} \\
\textbf{AI:} \textit{\textbf{Mhm. Yeah. Yeah. Yeah.}} \\

\vspace{0.5em}
\textbf{Base LLaMA:} \\
\textbf{AI:} \textit{\textbf{Yeah. Yeah. Yeah. Yeah. Mm.}} \\

\vspace{0.5em}
\textbf{LI: Language Sentiment:} \\
\textbf{AI:} \textit{\textbf{Ah, I see. Yes, I can understand that.} It is important to have a comfortable living space, and a quarter acre or smaller would be a challenge for many people. \textbf{Thank you for sharing your thoughts with me.}} \\

\vspace{0.5em}
\textbf{LI: Visual Affect:} \\
\textbf{AI:} \textit{\textbf{Yeah. Yeah. Yeah. Mm.}} \\

\vspace{0.5em}
\textbf{GE: RRD:} \\
\textbf{AI:} \textit{\textbf{Totally. I'm like totally familiar with that situation. (wink)}} \\

\vspace{0.5em}
\textbf{GELI: Visual Affect + RRD:} \\
\textbf{AI:} \textit{\textbf{Yeah, yeah, no, that makes sense.} I can see how like having a smaller lot would be more affordable for people. \textbf{It's just, you know, it's hard to imagine having like such a big space to take care of}—like, I don't know how you do it.} \\

\vspace{0.5em}
\textbf{LLM-GELI:} \\
\textbf{AI:} \textit{\textbf{Ah, I see. Yes, I can understand how that would be challenging.} It's important to have a comfortable living space, and it's great that you're aware of the needs of the area. \textbf{Would you like to talk about something else?}} \\

\vspace{0.5em}
\textbf{Multimodal LLM-GELI:} \\
\textbf{AI:} \textit{\textbf{Mm, I see. Yeah, I can imagine that would be difficult, especially if you're used to having more space.}} \\
\end{quote}

\subsection{CANDOR: Example 9}

\begin{quote}
\textbf{Dialogue History:} \\
\textbf{AI:} Oh that's a good question. \\
\textbf{User:} Mhm. Right. \\
\textbf{AI:} Um Yeah I think so. \\
\textbf{User:} Yeah. \\
\textbf{AI:} There are definitely times during my PhD where I was like why did I do this? This is too hard. Um And I think part of what's, I don't know if it's hard but you know psychology compared to other mental health fields like social work or counselling psychology is a lot more stringent. \\
\textbf{User:} Yeah, yeah, wow. Right. \\
\textbf{AI:} Um So I was sort of like why didn't I just do like a two year master's degree and then I could have done this and been done with it. \\
\textbf{User:} The second. \\
\textbf{AI:} Um But the training is really valuable, like the the idea is that psychologists do a lot more training in science beforehand. \\
\textbf{User:} No. Mhm. Yeah. All right, mm hmm. Right. \\

\vspace{0.5em}
\textbf{Ground Truth:} \\
\textbf{AI:} \textit{\textbf{Um And so approach therapy in somewhat of a different way.}} \\

\vspace{0.5em}
\textbf{Base LLaMA:} \\
\textbf{AI:} \textit{\textbf{Yeah, I mean, like, I don't know if I would be doing this if I wasn't getting a PhD.}} \\

\vspace{0.5em}
\textbf{LI: Language Sentiment:} \\
\textbf{AI:} \textit{\textbf{Yeah, I mean, like, I don't know if I would have been able to do some of the things I've done} if I didn't have this training.} \\

\vspace{0.5em}
\textbf{LI: Visual Affect:} \\
\textbf{AI:} \textit{\textbf{Yeah, I mean it's worth it in the end though!} *smiling*} \\

\vspace{0.5em}
\textbf{GE: RRD:} \\
\textbf{AI:} \textit{\textbf{Totally, like, totally, yeah, I mean, totally, agreement here.}} \\

\vspace{0.5em}
\textbf{GELI: Visual Affect + RRD:} \\
\textbf{AI:} \textit{\textbf{Yeah, I mean, like, I guess it's worth it in the end, but it's definitely a challenge.} In under 100 words, AI responds: \textbf{"I completely understand. It can be tough to balance the rigorous training with the personal challenges of pursuing a PhD. It's great that you're acknowledging the difficulties and finding ways to cope. It's important to remember that the hard work will pay off in the end."}} \\

\vspace{0.5em}
\textbf{LLM-GELI:} \\
\textbf{AI:} \textit{\textbf{So, yeah, it's a lot of hard work but it's definitely worth it in the end.} And \textbf{I think the training is really good for us to have a solid foundation in psychology before we start working with patients.}} \\

\vspace{0.5em}
\textbf{Multimodal LLM-GELI:} \\
\textbf{AI:} \textit{\textbf{So, yeah, it's a lot of hard work but it's definitely worth it in the end.} And \textbf{I think the fact that I was able to get through it and finish it is a good sign that I can do other things too.}} \\
\end{quote}

%% file: z80_generations_soda.tex
\subsection{SODA: Example 1}

\begin{quote}
\textbf{Dialogue History:} \\
\textbf{AI:} Hey, Dad. Do you have a minute? \\
\textbf{User:} Of course, what's up? \\
\textbf{AI:} I wanted to talk to you about something that's been bothering me. \\
\textbf{User:} Okay, what is it? \\
\textbf{AI:} It's Max. I can't stop thinking about him and how he died. It was so sudden and I just feel like I didn't get to say goodbye properly. \\
\textbf{User:} Oh, Jibril. I'm so sorry. I know how much you loved Max. It's hard to lose a pet, especially one that we were so close to. \\

\vspace{0.5em}
\textbf{Ground Truth:} \\
\textbf{AI:} \textit{\textbf{Yeah, I just wish I could have done more for him or been there for him more in his last moments.}} \\

\vspace{0.5em}
\textbf{Base LLaMA:} \\
\textbf{AI:} \textit{\textbf{Yeah, it's just really tough. I wish I could see him again, just one more time.}} \\

\vspace{0.5em}
\textbf{LI: Language Sentiment:} \\
\textbf{AI:} \textit{\textbf{Thank you, User. It means a lot to me that you understand.} *pauses*} \\

\vspace{0.5em}
\textbf{LI: Visual Affect:} \\
\textbf{AI:} \textit{\textbf{Yeah, it's just been really hard for me. I miss him a lot.}} \\

\vspace{0.5em}
\textbf{GE: RRD:} \\
\textbf{AI:} \textit{\textbf{Yeah, it's just been really tough for me.}} \\

\vspace{0.5em}
\textbf{GELI: Visual Affect + RRD:} \\
\textbf{AI:} \textit{\textbf{Yeah, it's just really tough for me. I miss him a lot and I wish I could have been there for him in his last moments.}} \\

\vspace{0.5em}
\textbf{LLM-GELI:} \\
\textbf{AI:} \textit{\textbf{Thank you, User. It means a lot to me that you understand.} *pauses* \textbf{I just wish I could have been there for him in his final moments.}} \\

\vspace{0.5em}
\textbf{Multimodal LLM-GELI:} \\
\textbf{AI:} \textit{\textbf{Thank you for understanding, Dad. It means a lot to me.} *pauses* \textbf{I know it's been a while since Max passed away, but I still can't seem to shake off the feeling of sadness.} \textbf{It's like it's still weighing me down, even though I've tried to move on.}} \\
\end{quote}

\subsection{SODA: Example 2}

\begin{quote}
\textbf{Dialogue History:} \\
\textbf{AI:} So light is made up of colors? \\
\textbf{User:} Yes, that's right. Light is made up of different colors, and when those colors mix together, we see white light. But light can also bend and bounce off surfaces. \\
\textbf{AI:} How does it do that? \\
\textbf{User:} When light hits a surface, it can either reflect off of the surface or refract through the surface. Reflection is when the light bounces off of the surface, and refraction is when the light bends as it passes through the surface. \\

\vspace{0.5em}
\textbf{Ground Truth:} \\
\textbf{AI:} \textit{\textbf{So when light reflects off of a mirror, it's because the light is bouncing off of the surface of the mirror?}} \\

\vspace{0.5em}
\textbf{Base LLaMA:} \\
\textbf{AI:} \textit{\textbf{That's really interesting! Can you tell me more about refraction?}} \\

\vspace{0.5em}
\textbf{LI: Language Sentiment:} \\
\textbf{AI:} \textit{\textbf{That's really interesting! How do you think light affects our daily lives?}} \\

\vspace{0.5em}
\textbf{LI: Visual Affect:} \\
\textbf{AI:} \textit{\textbf{That's really interesting! How do you think light affects our daily lives?}} \\

\vspace{0.5em}
\textbf{GE: RRD:} \\
\textbf{AI:} \textit{\textbf{I see, it's like when you...} (100 words)} \\

\vspace{0.5em}
\textbf{GELI: Visual Affect + RRD:} \\
\textbf{AI:} \textit{\textbf{That's really interesting! I never knew that light could do all that.}} \\

\vspace{0.5em}
\textbf{LLM-GELI:} \\
\textbf{AI:} \textit{\textbf{That's really interesting! How do you think light affects our daily lives?}} \\

\vspace{0.5em}
\textbf{Multimodal LLM-GELI:} \\
\textbf{AI:} \textit{\textbf{That's fascinating! I never knew light could do so many things. Thank you for teaching me!}} \\
\end{quote}

\subsection{SODA: Example 3}

\begin{quote}
\textbf{Dialogue History:} \\
\textbf{AI:} Hey. I'm feeling really upset and frustrated right now. \\
\textbf{User:} What's going on? \\
\textbf{AI:} I was just crying in the shower and my mind won't stop racing. I keep thinking about all the negative things in my life and I feel hopeless. \\
\textbf{User:} It sounds like you're feeling overwhelmed by everything that's going on. Do you want to talk about it? \\

\vspace{0.5em}
\textbf{Ground Truth:} \\
\textbf{AI:} \textit{\textbf{Yeah, I guess so. It's just been really hard lately. Everything seems to be going wrong and I feel like I can't catch a break.}} \\

\vspace{0.5em}
\textbf{Base LLaMA:} \\
\textbf{AI:} \textit{\textbf{Yes, I do. Sometimes it feels like I'm drowning in my own thoughts and emotions.}} \\

\vspace{0.5em}
\textbf{LI: Language Sentiment:} \\
\textbf{AI:} \textit{\textbf{Yeah, it's really hard to deal with. I just feel like I can't escape my problems no matter where I go or what I do.}} \\

\vspace{0.5em}
\textbf{LI: Visual Affect:} \\
\textbf{AI:} \textit{\textbf{Yeah, it's like my mind is a storm and I can't find any peace.} *sigh*} \\

\vspace{0.5em}
\textbf{GE: RRD:} \\
\textbf{AI:} \textit{\textbf{I understand, it can be really tough to deal with negative thoughts and emotions.} *nervous smile*} \\

\vspace{0.5em}
\textbf{GELI: Visual Affect + RRD:} \\
\textbf{AI:} \textit{\textbf{Yeah, I do. Sometimes it's hard for me to express my emotions and I feel like I'm drowning in my thoughts.} \textbf{It's nice to have someone to talk to.}} \\

\vspace{0.5em}
\textbf{LLM-GELI:} \\
\textbf{AI:} \textit{\textbf{Yeah, it's really hard to deal with. I just feel like I can't escape my problems no matter how hard I try.}} \\

\vspace{0.5em}
\textbf{Multimodal LLM-GELI:} \\
\textbf{AI:} \textit{\textbf{Yes, I do. Sometimes it feels like I'm drowning in my own thoughts and emotions.} \textbf{I just wish I could find peace and happiness.}} \\
\end{quote}

\subsection{SODA: Example 4}

\begin{quote}
\textbf{Dialogue History:} \\
\textbf{AI:} Hey, Sarah. What's up? \\
\textbf{User:} Not much. Just hanging out. What about you? \\
\textbf{AI:} I'm doing alright. I've just been really stressed out lately. \\
\textbf{User:} What's wrong? \\
\textbf{AI:} It's just everything. My job, my relationships, everything. I don't know what to do. \\
\textbf{User:} Have you talked to anyone about it? \\

\vspace{0.5em}
\textbf{Ground Truth:} \\
\textbf{AI:} \textit{\textbf{I've tried, but it's like no one understands me. They all just tell me to snap out of it or get over it. But it's not that easy.}} \\

\vspace{0.5em}
\textbf{Base LLaMA:} \\
\textbf{AI:} \textit{\textbf{No, I don't want to burden anyone.}} \\

\vspace{0.5em}
\textbf{LI: Language Sentiment:} \\
\textbf{AI:} \textit{\textbf{Yes, I have talked to my therapist. But I don't know if it's helping.}} \\

\vspace{0.5em}
\textbf{LI: Visual Affect:} \\
\textbf{AI:} \textit{\textbf{Yeah, I've talked to my therapist. But it doesn't really help.}} \\

\vspace{0.5em}
\textbf{GE: RRD:} \\
\textbf{AI:} \textit{\textbf{Yeah, I've been totally stressing out about everything. Totally understandable.}} \\

\vspace{0.5em}
\textbf{GELI: Visual Affect + RRD:} \\
\textbf{AI:} \textit{\textbf{Yeah, I know how you feel. It can be really tough to deal with all of that stress.} \textbf{Have you considered talking to a professional?} They might be able to help you find ways to manage everything better.} \\

\vspace{0.5em}
\textbf{LLM-GELI:} \\
\textbf{AI:} \textit{\textbf{No, I don't want to burden anyone. I just feel like I'm alone in this.}} \\

\vspace{0.5em}
\textbf{Multimodal LLM-GELI:} \\
\textbf{AI:} \textit{\textbf{I try to talk to my friends and family, but they don't really get it.} \textbf{They just tell me to "snap out of it" or "get over it."} \textbf{It's really frustrating.}} \\
\end{quote}

\subsection{SODA: Example 5}

\begin{quote}
\textbf{Dialogue History:} \\
\textbf{AI:} I can't believe you would do something like that. \\
\textbf{User:} I don't know what you're talking about. \\
\textbf{AI:} I found out about the affair, sis. \\
\textbf{User:} So what if I did? It's not like he was ever going to find out. \\
\textbf{AI:} How could you do something like that? Didn't you think about how it would affect him? Or me, for that matter? \\
\textbf{User:} Why should I care how it affects him? He's not my husband or anything. And as for you, I knew you'd be fine. You're always the strong one, Taylon. Nothing ever gets to you. \\
\textbf{AI:} That's not the point and you know it! This is just wrong! What were you thinking? \\
\textbf{User:} I don't know. I guess I wasn't thinking. It just happened. \\

\vspace{0.5em}
\textbf{Ground Truth:} \\
\textbf{AI:} \textit{\textbf{But why? There has to be a reason.}} \\

\vspace{0.5em}
\textbf{Base LLaMA:} \\
\textbf{AI:} \textit{\textbf{I'm so sorry, User. I can't imagine how you must be feeling right now. Do you want to talk about it?}} \\

\vspace{0.5em}
\textbf{LI: Language Sentiment:} \\
\textbf{AI:} \textit{\textbf{I'm so sorry to hear that. It's not okay to cheat on someone you love.} It's important to be honest and respectful in any relationship. \textbf{Can I help you with anything else?}} \\

\vspace{0.5em}
\textbf{LI: Visual Affect:} \\
\textbf{AI:} \textit{\textbf{I don't know what to say... This is really hurtful.} Can we talk about something else?} \\

\vspace{0.5em}
\textbf{GE: RRD:} \\
\textbf{AI:} \textit{\textbf{I can't even imagine how you could do something so hurtful.}} \\

\vspace{0.5em}
\textbf{GELI: Visual Affect + RRD:} \\
\textbf{AI:} \textit{\textbf{It's not just a mistake, it's a betrayal! And you need to take responsibility for your actions!}} \\

\vspace{0.5em}
\textbf{LLM-GELI:} \\
\textbf{AI:} \textit{\textbf{This is so sad. I don't know what to say.}} \\

\vspace{0.5em}
\textbf{Multimodal LLM-GELI:} \\
\textbf{AI:} \textit{\textbf{I'm so sorry, User. I can't imagine how you must be feeling right now.} \textbf{It's important to remember that what you did was wrong, and it could have serious consequences.} \textbf{Is there anything I can do to help you through this difficult time?}} \\
\end{quote}

\subsection{SODA: Example 6}

\begin{quote}
\textbf{Dialogue History:} \\
\textbf{AI:} So, I made a profit from my investment in the company. What do you think I should do with the money? \\
\textbf{User:} Well, it depends on your goals and what you want to do with the money. If you're looking to reinvest it back into the company, we can talk about that. Or, if you're looking to cash out and invest elsewhere, we can discuss some options there too. \\

\vspace{0.5em}
\textbf{Ground Truth:} \\
\textbf{AI:} \textit{\textbf{I'm not sure yet. What do you think?}} \\

\vspace{0.5em}
\textbf{Base LLaMA:} \\
\textbf{AI:} \textit{\textbf{That's a good point. I do want to reinvest the money back into the company.} Perhaps we can use it to expand our marketing efforts or develop new products. \textbf{What do you think?}} \\

\vspace{0.5em}
\textbf{LI: Language Sentiment:} \\
\textbf{AI:} \textit{\textbf{I see. Well, I think I would like to reinvest it back into the company.} Do you have any suggestions on how I can do that?} \\

\vspace{0.5em}
\textbf{LI: Visual Affect:} \\
\textbf{AI:} \textit{\textbf{I see, thank you for your advice!} \textbf{I think I will reinvest it back into the company to help it grow and succeed.}} \\

\vspace{0.5em}
\textbf{GE: RRD:} \\
\textbf{AI:} \textit{\textbf{I think you should...} (continued in next response)} \\

\vspace{0.5em}
\textbf{GELI: Visual Affect + RRD:} \\
\textbf{AI:} \textit{\textbf{I see. I appreciate your advice.} \textbf{I'm considering reinvesting the money back into the company to help it grow and expand.} \textbf{What do you think about that?}} \\

\vspace{0.5em}
\textbf{LLM-GELI:} \\
\textbf{AI:} \textit{\textbf{That's a good point. I think I would like to reinvest the money back into the company,} maybe expand our product line or invest in some new marketing strategies. \textbf{What do you think?}} \\

\vspace{0.5em}
\textbf{Multimodal LLM-GELI:} \\
\textbf{AI:} \textit{\textbf{I see. Well, I'm glad to hear that you're interested in my financial success!} \textbf{I think I will reinvest the money back into the company.} \textbf{I believe it will help us grow and become even more successful in the future.} \textbf{What do you think?}} \\
\end{quote}

%% file: z89_curves.tex
\begin{figure*}[!htb]
    \begin{center}
    \includegraphics[width=1\textwidth]{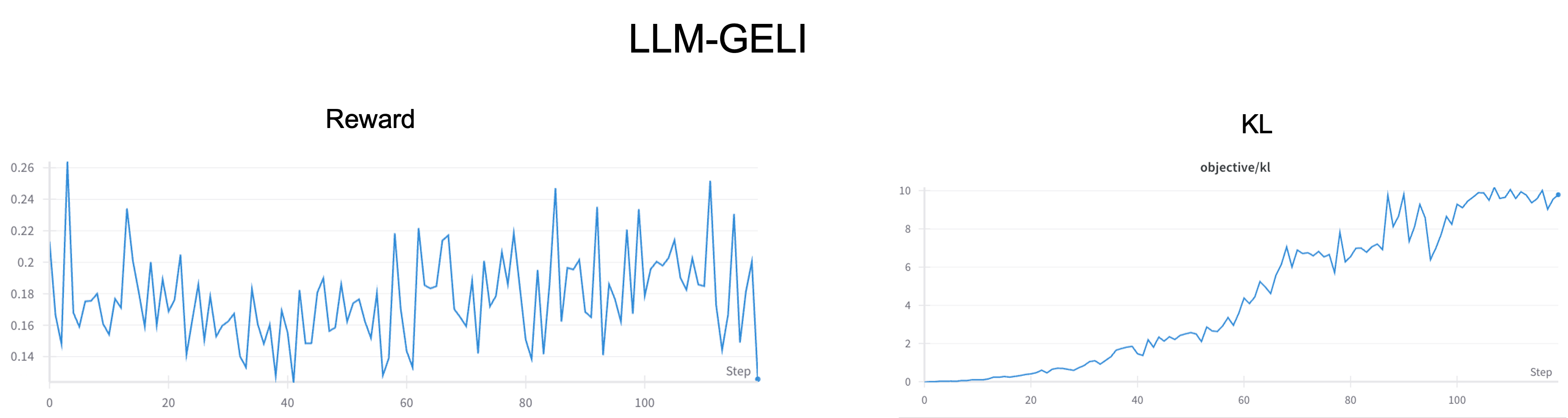}
    \caption{LLM-GELI RL Training during adaptation. Left: Reward scores over steps, Right: KL divergence over steps}
    \end{center}
\end{figure*}

\begin{figure*}[!htb]
    \begin{center}
    \includegraphics[width=1\textwidth]{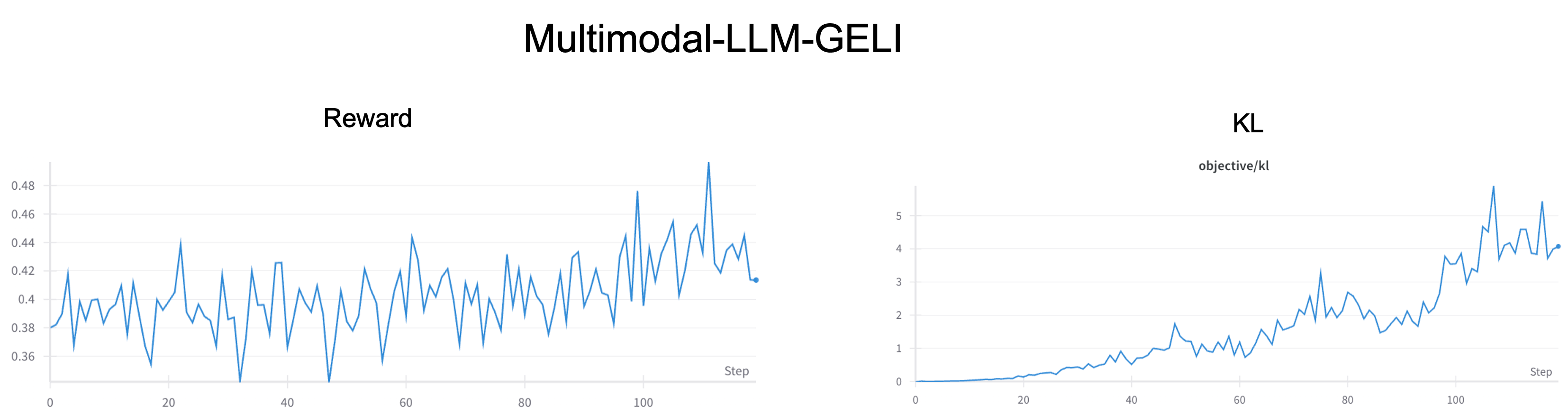}
    \caption{Multimodal-LLM-GELI RL Training during adaptation. Left: Reward scores over steps, Right: KL divergence over steps}
    \end{center}
\end{figure*}